%% file: main.tex
\pdfoutput=1
\documentclass[11pt]{article}

\usepackage[]{acl}
\usepackage{booktabs}
\usepackage{enumitem}
\usepackage{tcolorbox}
\usepackage{bm}
\usepackage{multirow}
\usepackage{stfloats}
\usepackage{float}
\usepackage{amsmath}
\DeclareMathOperator*{\argmax}{arg\,max}
\usepackage{makecell}
\usepackage{graphicx}
\usepackage{longtable}
\usepackage{times}
\usepackage{latexsym}
\usepackage[ruled,noend]{algorithm2e}
\usepackage{subcaption}
\usepackage[T1]{fontenc}

\newcommand{\name}{\textsf{$\textbf{C}^{3}$}}

\usepackage{setspace}
\AtBeginDocument{
  \addtolength\abovedisplayskip{-0.05\baselineskip}%
  \addtolength\belowdisplayskip{-0.05\baselineskip}%
  \addtolength\abovedisplayshortskip{-0.05\baselineskip}%
  \addtolength\belowdisplayshortskip{-0.05\baselineskip}%
}
\usepackage{titlesec}
\titlespacing*{\section}{0pt}{0.25\baselineskip}{0.25\baselineskip}
\titlespacing*{\subsection}{0pt}{0.25\baselineskip}{0.25\baselineskip}
\titlespacing*{\subsubsection}{0pt}{0.25\baselineskip}{0.25\baselineskip}

\usepackage[utf8]{inputenc}
\usepackage{microtype, colortbl, xcolor}

\usepackage{inconsolata}
\definecolor{LightCyan}{rgb}{0.88,1,1}
\definecolor{LightRed}{rgb}{1,0.9,0.8}
\definecolor{LightYellow}{rgb}{1,1,0.8}
\title{{\name}: Confidence Calibration Model Cascade for Inference-Efficient Cross-Lingual Natural Language Understanding}

\author{
  Taixi Lu$^{1}$\thanks{\;\;equal contribution}, Haoyu Wang$^{2}$\footnotemark[1], Huajie Shao$^{3}$, Jing Gao$^{2}$, Huaxiu Yao$^{1}$ \\
  $^{1}$UNC-Chapel Hill, $^{2}$Purdue University, $^{3}$College of William and Mary\\
  \tt{taixi@email.unc.edu, wang5346@purdue.edu, huaxiu@cs.unc.edu}
 }

\begin{document}
\maketitle
\input{latex/sections/abstract}

\input{latex/sections/introduction}

\input{latex/sections/relatedwork}

\input{latex/sections/preliminary}
\input{latex/sections/methodology}

\input{latex/sections/experiment}

\input{latex/sections/analysis}
\input{latex/sections/hyper}
\input{latex/sections/conclusion}

\input{latex/sections/limitation}
\bibliography{anthology,custom}

\input{latex/sections/appendix}

\end{document}

%% file: latex/sections/abstract.tex
\begin{abstract}

Cross-lingual natural language understanding (NLU) is a critical task in natural language processing (NLP). Recent advancements have seen multilingual pre-trained language models (mPLMs) significantly enhance the performance of these tasks. However, mPLMs necessitate substantial resources and incur high computational costs during inference, posing challenges for deployment in real-world and real-time systems. Existing model cascade methods seek to enhance inference efficiency by greedily selecting the lightest model capable of processing the current input from a variety of models, based on model confidence scores. Nonetheless, deep models tend to exhibit overconfidence, and confidence distributions vary across languages. This leads to the emission of confident but incorrect predictions by smaller models, hindering their ability to generalize effectively across test languages. In this study, we introduce \textbf{C}onfidence \textbf{C}alibration \textbf{C}ascade ({\name}), a simple yet effective method that involves calibration prior to cascade inference, thereby enhancing cascade accuracy through more reliable predictions. Our evaluation of {\name}, using both encoder-only and decoder-only language models, across five cross-lingual benchmarks covering classification and generation tasks, shows that {\name} markedly surpasses all leading baselines.
\end{abstract}

%% file: latex/sections/introduction.tex
\section{Introduction}

Pre-trained language models (PLMs), such as BERT \cite{devlin-etal-2019-bert}, RoBERTa \cite{liu2019roberta}, T5 \cite{2020t5}, and Llama \cite{touvron2023llama}, have exhibited remarkable performance across various natural language processing (NLP) tasks. Notably, their multilingual versions have demonstrated impressive zero-shot transfer capabilities in cross-lingual settings \cite{pires-etal-2019-multilingual-bert,xlm-roberta}. In these scenarios, PLMs are fine-tuned on English data, often with limited or even without data from other languages, yet they acquire the proficiency to handle tasks in different languages. However, multilingual PLMs are typically constructed using stacked transformer layers or their variants, employing self-attention mechanisms to capture diverse and distant dependencies among tokens. The use of self-attention introduces significant computational complexity. Consequently, the inference complexity of multilingual PLMs has become a bottleneck, limiting their deployment on devices sensitive to latency and constrained by computational resources.
\begin{figure}[h!]
    \centering
    \hfill
    \includegraphics[width=1.0\linewidth]{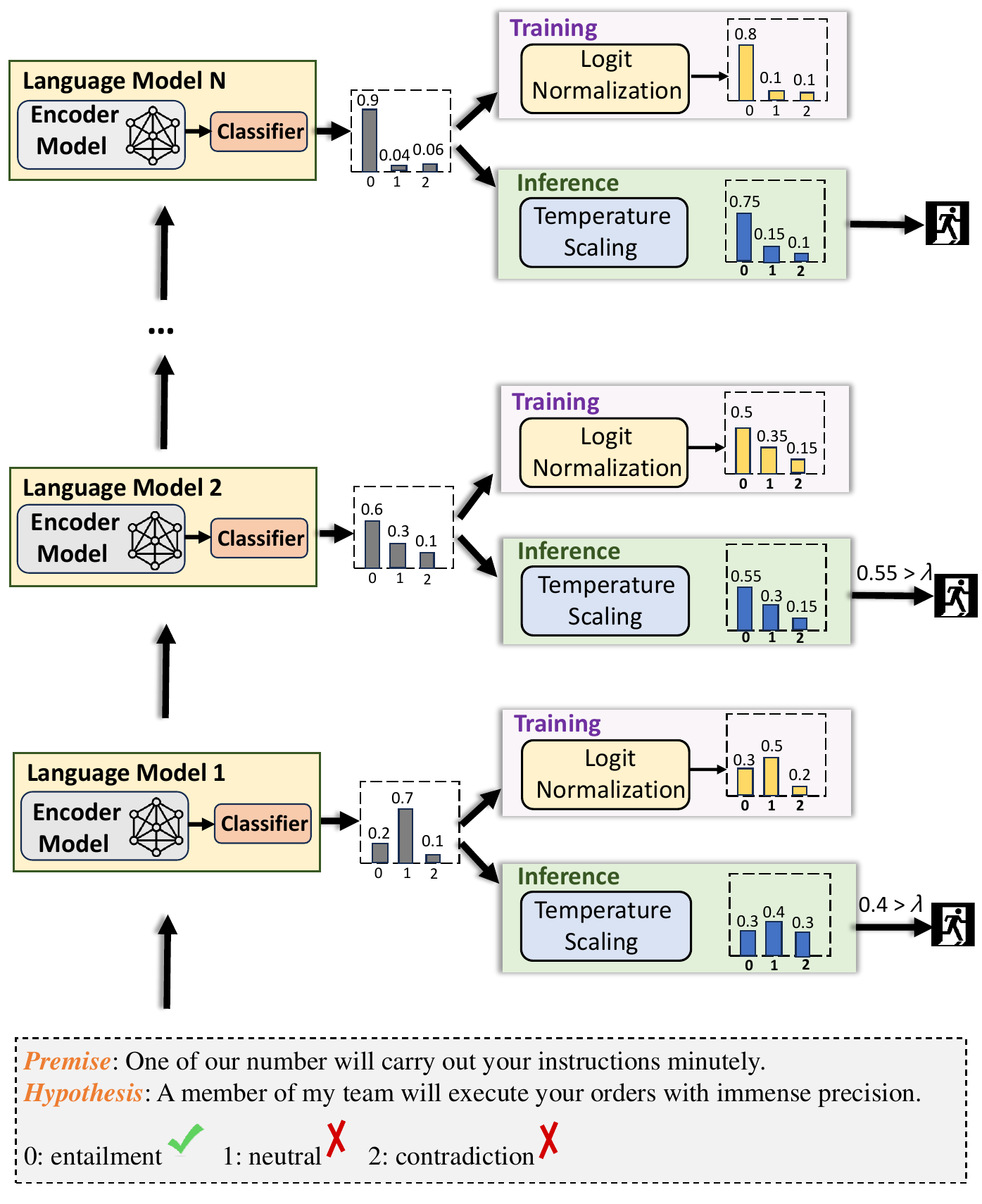}
    \caption{An illustration of our {\name} framework (for classification task) for speeding up natural language inference yet retain the most accuracy, especially in OOD data. We leverage Logit Normalization at training time and Temperature Scaling at inference time to calibrate each model so that the model will yield more reliable confidence score for cascade decisions. For Large Language Model (e.g., GPT-4, Llama) inference where there is no training involved, we simply remove the training module. The $\lambda$ represents the confidence score.}
    \label{fig:framework}
    \vspace{-1.5em}
\end{figure}

To fulfill the stringent requirements for efficient inference in applications, various methods have been proposed to accelerate Pre-trained Language Model (PLM) inference. These methods include model compression \cite{sanh2019distilbert, jiao-etal-2020-tinybert, sun-etal-2020-mobilebert, pkd}, early exiting \cite{deebert, pabee, global}, and model cascading \cite{cascadebert, wisdom}. Among these, model cascading methods are particularly appealing for several reasons: 1)~They do not depend on specific hardware support, such as custom chips and GPUs. 2)~They eliminate the need to train an inference-efficient model from scratch on the pre-training corpora. 3)~They offer flexibility to adapt to the latest, incrementally powerful PLMs. Model cascading methods involve the aggregation of multiple PLMs with different sizes. Confidence scores are computed sequentially, ranging from small to large size models, to determine the appropriate model to employ. Once a confidence score surpasses a threshold, the corresponding model is selected, and the inference process ends.

Cascade-based models, however, exhibit notable limitations in cross-lingual scenarios. The confidence score, which measures the probability of the current prediction being correct in cascade-based models, is determined by the maximum output probability, the mean of the output probability, or the entropy of the output probability. Unfortunately, neural networks often generate unreliable confidence scores, particularly in out-of-distribution (OOD) scenarios \cite{calibration, logitnorm, han2024selective,choi2023conservative,liang2022holistic}. The cross-lingual task represents a typical OOD setting, where the training data exhibits significantly different distributions compared to the testing data. Consequently, the distribution of confidence scores on the training data differs from that on the testing data. Applying a threshold derived from the training set, based on pre-defined inference budgets, may prove either too conservative or radical for the testing data, leading to a performance or efficiency drop.

To tackle the challenges outlined earlier, we introduce a \textbf{C}onfidence \textbf{C}alibration \textbf{C}ascade ({\name}) framework for efficient cross-lingual inference. The motivation behind our proposed approach is to calibrate the confidence of Multilingual Pre-trained Language Models (mPLMs), allowing the threshold determined on English data to be applicable to other languages. Specifically, we introduce a plug-in calibration step at the base of mPLMs. Initially, we normalize the logits to alleviate over-confidence during model fine-tuning. Subsequently, we implement a temperature scaling step to adjust the logits with a learnable scalar parameter. The proposed framework calibrates each individual model in the cascade, providing more reliable confidence scores. This, in turn, enhances the model's performance and generalization capabilities, leading to consistent improvements in efficiency and accuracy across different languages. Importantly, the proposed framework only requires an extra calibration module at the base of mPLMs, preserving the original architectures of mPLMs. Hence, it demonstrates flexibility to accommodate the latest models with minimal additional training overhead.

The primary contributions of this paper is {\name}, a flexible and effective framework for enhancing efficiency in cross-lingual inference. To the best of our knowledge, this is the first work dedicated to the design of inference-efficient models specifically tailored for cross-lingual scenarios. Based on the observation of a notable overconfidence phenomenon in both encoder-only PLMs and decoder-only PLMs in cross-lingual scenarios, and considering that the extent of overconfidence appears to be correlated with linguistic distance, we introduce a plug-in calibration module to address this issue. Extensive experiments on five cross-lingual benchmarks, including XNLI, PAWS-X, QAM, GSM8k, and TabMWP, where the first three are text classification datasets and the later two are generation datasets, indicate that the proposed {\name} outperforms baselines significantly and achieves a good efficiency-accuracy trade-off, e.g., preserving 98.10\% of BERT’s performance and 95.28\% of Llama-2's performance on classification task and an average of 74.3\% performance on generation task only half the computation costs.

%% file: latex/sections/relatedwork.tex
\section{Related Work}
\subsection{Inference Acceleration for PLMs}
Existing approaches for accelerating the inference of pre-trained language models~(PLMs) can be categorized into two types: 1)~model compression-based methods and 2)~dynamic network-based methods. The proposed method is more relevant to dynamic network-based methods, which are discussed in more detail as follows. The discussion about the model compression-based methods can be found in the Appendix \ref{sec:appendix_related_work}.

Dynamic network-based methods leverage models with different sizes based on input data when inference. Early existing methods and cascade methods are two typical dynamic networks. 

\noindent\textbf{Early exiting Methods.} There are a number of early exiting methods that accelerate model inference by emitting predictions from an inner layer. FastBERT \cite{weijie2020fastbert}, DeeBERT \cite{deebert}, and SDN \cite{kaya2019shallow}, are score-based methods, using the entropy of the prediction probability and the maximum of the predicted distribution as the score for exit decision. BERxiT \cite{xin2021berxit} is another type of early-exiting method that learns to exit. Also, patience-based methods, such as PABEE \cite{pabee}, SENTEE \cite{li2021accelerating}, and LeeBERT \cite{zhu2021leebert}, ensemble scores from multiple layers to make decision. However, the problem with early existing methods is that the model loses high-level semantic understanding without going into the higher layers. 

\noindent\textbf{Cascade Methods.} The main difference between cascade and early existing methods is cascade methods choose models with different sizes to make predictions while early existing methods choose a layer to exit with respect to a specific model. Specifically, cascades combine different sizes of models and go through each model sequentially from the smallest one to the largest one. Once the prediction from one model is confident enough, the prediction is emitted. Many previous works incorporate model cascade for faster inference on certain tasks. For example, \citet{viola2001rapid} built a cascade of classifiers with increasing complexity to speed up facial recognition; CascadeBERT ensembles two models prediction and regularizes them with a difficulty-awareness regularization \cite{cascadebert}; Window-Based Cascade caches output from multiple levels and compare them to a window threshold \cite{Xia_2023_window}.
\subsection{Model Calibration}
As models become larger, they also tend to be less calibrated, meaning the confidence output by the model does not reflect the true probability. In fact, large models are found to be easily overconfident in a lot of scenarios. We found that the problem of miscalibration is essentially important in cascades, where confidence is directly used as a metric to determine whether the prediction from the smaller models are reliable. There are various calibration methods in previous works~\cite{tian2023just}. For example, Temperature Scaling divides the logits by a learned parameter from the validation set to calibrate the model after training \cite{calibration}; Logit Normalization incorporates a modified loss function to learn to produce logits with smaller norms \cite{logitnorm}; AugMax uses data augmentation techniques to calibrate models \cite{wang2021augmax}.

%% file: latex/sections/preliminary.tex
\section{Preliminaries: Model Cascade}

In this section, we introduce the overall flow of the vanilla model cascade. The model cascade consists of $n$ PLMs, denoted as $\{M_{i}\}_{i=1}^{n}$, with the sizes of these PLMs arranged in ascending order, i.e. $|M_{i}|<|M_{j}|$ for $i<j$ where $|\cdot|$ represents the size of the model. The cascade model leverages each PLM for inference sequentially and computes the confidence score based on prediction logits of current PLM. If the confidence score exceeds a predefined threshold, the cascade model terminates its inference. Specifically, for an input text sequence $t$ and the current PLM $M_{i}$, the output logits can be represented as
\begin{align}
    L=M_{i}(t),
\end{align}
where $L=\{l_{j}\}_{j=1}^{q}$ and $q$ is the number of classes in the label. Subsequently, it computes the confidence score based on the logits $L$, such as by using averaged probability or maximum probability. We illustrate the calculation using maximum probability as an example:
\begin{align}
    C=\argmax_{j}\frac{\exp{l_{j}}}{\sum_{k}\exp(l_{k})}.
\end{align}
If it has traveled all models or $C>\lambda$, where $\lambda$ is a threshold, the inference ends and returns the current logits $L$.

%% file: latex/sections/methodology.tex
\section{Confidence Calibration Cascade}

Given $n$ pre-trained language models~(PLMs) with ascending sizes $\{M_{i}\}_{i=1}^{n}$ and source language~(English in our paper) training set $\mathcal{D}=\{(x^{i},y^{i})\}_{i=1}^{N}$, where $x^{i}$ is the input sequence, $y^{i}$ is the corresponding label and $N$ is the number of training data, the aim is to fine-tune these PLMs and select the most suitable PLM that enhances the inference efficiency while preserving accuracy on both source and target languages with respect to the largest PLM $M_{n}$.
\subsection{Overview}

To enhance model inference efficiency in cross-lingual scenarios, we propose a \textbf{c}onfidence \textbf{c}alibration model \textbf{c}ascade method~(\name), as illustrated in Fig.~\ref{fig:framework}. The motivation behind this approach is to select the most lightweight model for each input based on the model confidence scores. To ensure the reliability of the confidence score, we introduce an additional calibration module integrated into vanilla cascade methods, which includes logit normalization and temperature scaling. We present the proposed {\name} with encoder-only language models in Section~\ref{sec:encoder-only} and decoder-only language models in Section~\ref{sec:decoder-only}.

\subsection{Confidence Calibration for Language Model Cascade}
\label{sec:encoder-only}

In the vanilla model cascade method, a sample sequentially progresses through models of increasing size. A prediction is made when a model's confidence score exceeds a predetermined threshold, $\lambda$, or when the final model in the sequence is reached.

Our proposed {\name} approach to cascading language models introduces two additional steps to the standard model cascade procedure. These steps aim to align the confidence distributions across various languages and models of different sizes. The first step involves integrating a Logit Normalization layer during the fine-tuning of each language model. Consider $n$ language models $\{M_{1},..,M_{n}\}$. Each model is fine-tuned using a task-specific loss function 
$\ell(\bm{x},\bm{y})$, where the input logits are normalized. This can be expressed as:
\begin{align}
\ell(\bm{x},\bm{y})=-\log\frac{\exp(\frac{\bm{l}_y}{\tau\|\bm{l}\|})}{\sum_{i=1}^q\exp(\frac{l_i}{\tau\|\bm{l}\|})},
\end{align}
where $q$ is the number of classes, $\bm{l}=[\bm{l}_{1},...,\bm{l}_{q}]$ is the logits, $\|\bm{l}\|$ is the norm of the logits, and $\tau$ is a hyper-parameter to modulate the model's output logits. The rationale for Logit Normalization is to counteract the tendency of the cross-entropy loss function, which often encourages the model to produce increasingly large logits during training, leading to overconfident Softmax scores. By maintaining a constant 
$\ell_{2}$ norm of logits, Logit Normalization seeks to address this issue.

The second step involves applying temperature scaling to each fine-tuned model. Temperature scaling adjusts the model output logits by scaling them with a temperature parameter learned from the validation data. More precisely, we learn the temperature parameter $T_i$ for each model $M_i$, and then we scale the model logits $\bm{l}$ by the temperature when calculating the confidence score $C$.
\begin{align}
    C_i = \argmax_j \frac{\exp(\bm{l}_j/T_i)}{\sum_{k}^{ }{\exp(\bm{l}_k/T_i)}}\
    \label{equ3}
\end{align}

These enhancements aim to calibrate model predictions and enhance the reliability of confidence scores, a critical factor in model selection within a cascade framework.

\subsection{Confidence Calibration for Large Language Model Cascade}
\label{sec:decoder-only}

\begin{figure}[t]
    \centering
    \hfill
    \includegraphics[width=0.9\linewidth]{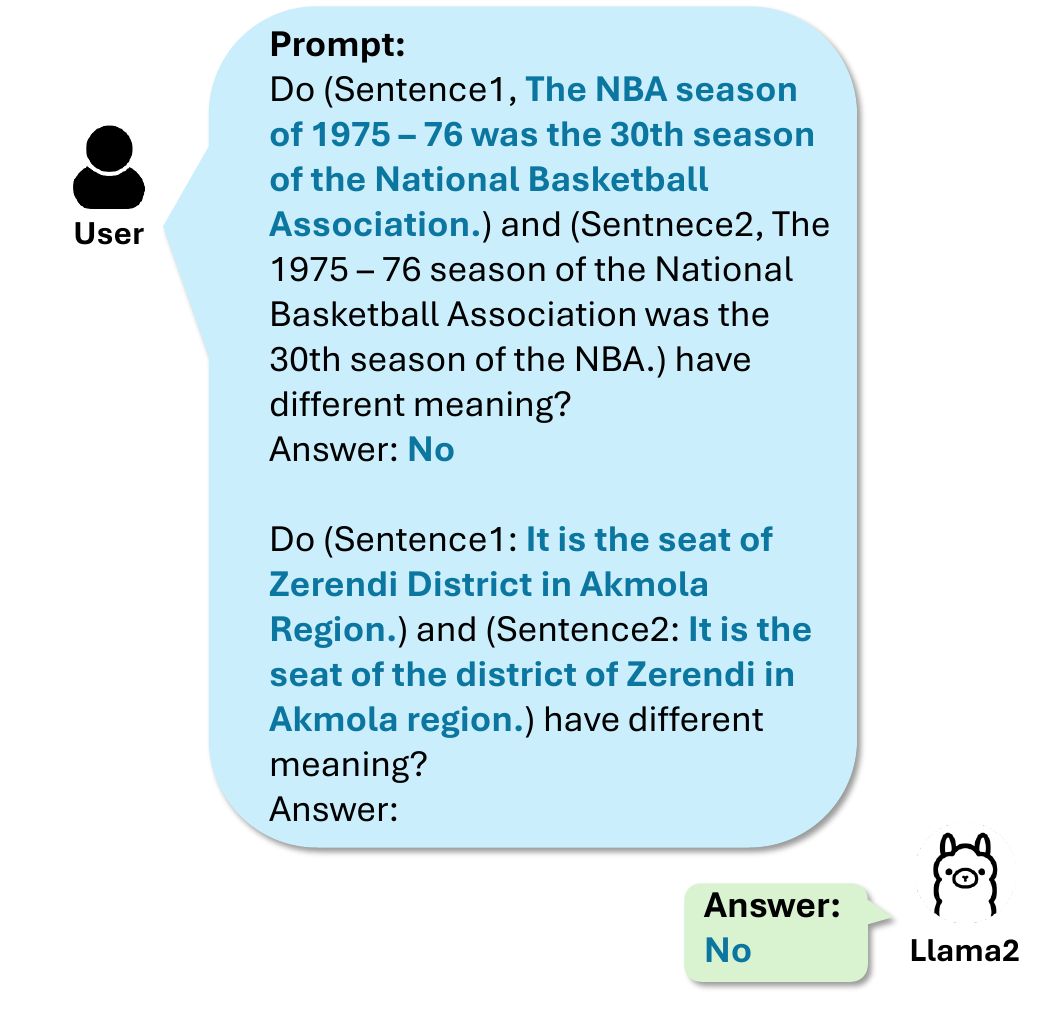}
    \caption{\textbf{Example {\name} one-shot prompt on Llama-2 for PAWS-X task.} The answer candidates set would be \{Yes, No\}, and the embedding set would be \{y, n\}, where y is the Llama embedding of the word "Yes," and n is the Llama embedding for the word "No."}
    \label{fig:conversation}
    \vspace{-1.5em}
\end{figure}

In this section, we illustrate the application of our proposed method {\name} to large language models (LLMs). As with {\name} for encoder-only LMs, we compile a collection of $n$ LLMs of varying sizes. However, in contrast to the full fine-tuning approach described in Section~\ref{sec:encoder-only}, the adaptation of LLMs to specific downstream tasks is more effectively achieved through zero-shot or few-shot in-context learning. This is due to the substantial number of parameters in LLMs and their inherent capacity for generalization. Following this approach, we incorporate the task description and examples in English into a prompt, as detailed in Figure \ref{fig:conversation}. The LLM $M_{i}$ then generates logits for the next token from the vocabulary, which can be represented as:
\begin{align}
    \bm{l}=\text{Logits}_{M_{i}}(\text{Prompt}).
\end{align}
For cross-lingual NLU tasks, particularly classification scenarios, we extract the logit corresponding to each class from $l$ using the token IDs for each class. This yields the logits for each class, denoted as $\{\bm{l}_{\bm{c}_{j}}\}_{j=1}^{q}$, where $\bm{c}_{j}$ is the $j$-th class. We then apply temperature scaling to calibrate each model's output using a temperature $T_i$ same as Section~\ref{sec:encoder-only} to achieve model confidence. Finally, we determine a threshold $\lambda$ for model selection during inference. The complete framework is outlined in Alg.~\ref{alg:1} in the appendix.

Furthermore, more generally, for generation tasks, unlike classification tasks where a set of logits per class can be obtained for temperature scaling, we employ token-level relevance~\cite{duan2023shifting} for calibration. Specifically, we compute the entropy for each token in the generated sequence, i.e. 
\begin{align}
    E(\bm{Z}_i)=-\log(p(\bm{Z}_i)),
\end{align} 
where $\bm{Z_}i$ is the $i$th token in the generated sequence. For each token $\bm{Z}_i$, we remove it from the sequence and utilize a pre-trained sentence similarity model to calculate the similarity, denoted as $\bm{R}_i$. The entropy of the entire sequence is determined by
\begin{align}
    E(\bm{Z})=\frac{1}{n}\sum_{i=0}^nE(\bm{Z}_i)(1-\bm{R}_i).
\end{align}
Finally, we establish a threshold, $\lambda$, for $\bm{Z}_i$ to manage the cascading of models effectively.

%% file: latex/sections/experiment.tex
\section{Experiment}

In this section, we evaluate the performance of {\name} on three cross-lingual benchmarks, aiming to answer the following questions: (1) How does {\name} perform compared to state-of-the-art baselines? (2) Is the proposed calibration method effective to reduce the calibration error? (3) How does the performance change with varing hyper-parameter $\tau$ in Logit Normalization? 

\subsection{Experimental Setup}

\input{tables/dataset}
\paragraph{Datasets} To evaluate the performance of our proposed {\name} on \textbf{classification} tasks, we conduct comprehensive experiments on three widely-used cross-lingual benchmarks: XNLI \cite{conneau2018xnli,wang2023macedon}, PAWS-X \cite{pawsx2019emnlp}, and QAM \cite{Liang2020XGLUEAN}. These datasets cover natural language inference, paraphrase identification, and question-answering matching tasks, respectively. Model fine-tuning is exclusively performed using the English training set, with subsequent evaluation conducted directly on testing data in other languages. Additionally, to evaluate the performance of our proposed {\name} on \textbf{generation} tasks, we also use GSM8k~\cite{cobbe2021gsm8k} and TabMWP~\cite{tabmwp} datasets. These two datasets both evaluate a model's mathematical reasoning capabilities, featuring questions as mathematical problems and answers in free-form text. Due to limited computational resources for running Large Language Models, we randomly selected 500 samples from the test set of each dataset for evaluation. Subsequently, we utilized the Google Translate API to translate these examples into five other languages: Spanish, German, Chinese, Japanese, and Korean. Detailed dataset statistics are presented in Table \ref{tab:dataset} in the appendix.

\paragraph{Baselines}
We adopt the following state-of-the-art methods as baselines: (1) \textbf{PABEE.} The early exiting method PABEE emits predictions from an inner layer of a model if the prediction remains consistent for a certain number of consecutive instances \cite{pabee}; (2) \textbf{DeeBERT}, which is another early exiting method that utilizes off-ramps layers to determine whether the prediction is confident enough at an inner layer \cite{deebert}; (3) \textbf{CasacdeBERT}, which is a cascade-based method that performs inference based on complete models with early stopping criteria and a difficulty-aware regularization \cite{cascadebert}; (4) \textbf{Cascade}, which is also a cascade-based method that goes through each model sequentially until the stopping restriction is met \cite{wisdom}. 

\subsection{Performance Comparison} 

In this section, we vary the threshold for each method and compare their accuracy on the test set, maintaining an identical speed-up ratio to answer RQ1. The speed-up ratio ($S$) is computed by dividing the number of Floating Point Operations (FLOPs) that the largest model (XLM-RoBERTa-large and Llama2-70b-chat-hf) would require by the number of FLOPs of the current method. Consistent with previous work, we also consider three distinct speed-up ratios, i.e. 2, 3, and 4, for a comprehensive comparison.

\input{tables/XNLI1} 
\input{tables/PAWSX}
\input{tables/QAM}
\subsubsection{Encoder-only Language Model}
We fine-tune encoder-only language models, including DistilBERT, mBERT-base, XLM-RoBERTa-base, and XLM-RoBERTa large, on the XNLI, PAWS-X, and QAM datasets, as illustrated in tables from Table \ref{tab:xnli1} to Table \ref{tab:qam}. According to these tables, the proposed method (\name) outperforms existing methods. Notably, {\name} exhibits superior performance compared to other cascade-based methods, especially in languages distant from English. For instance, on Thai, {\name} achieves an accuracy of 67.76\%, surpassing the baseline Cascade by 8.26\% at a speed-up ratio of 3.

\input{tables/decoder_result}
\subsubsection{Decoder-only Language Model}

We also conduct experiments using Llama2 as the backbone on PAWS-X to investigate the generalization capability of the proposed method ({\name}) for large-scale decoder-only language models on classification task. The results are presented in Table \ref{tab:llama}. Notably, the framework exhibits only a marginal 3\% decrease in accuracy when subjected to a two-fold speed-up, another approximately 3\% decrease for a three-fold speed-up, and less than an additional 2\% decrease for a four-fold speed-up. It's noteworthy that our findings lack accuracy data for the baseline Cascade. This absence is due to the remarkably concentrated confidence scores of Cascade, predominantly clustering around the value 1, resulting from the overconfident predictions by Llama-2. Consequently, establishing a threshold to achieve a specific speedup ratio becomes unfeasible, marking a limitation of Cascade.

We further present the results of our experiments using Llama2 as the backbone architecture on the GSM8k and TabMWP datasets in Table ~\ref{tab:llama}, demonstrating the generalizability of our method to generation tasks. In the majority of languages within these datasets, our method surpasses the Cascade method by margins ranging from 2\% to 6.2\%. In other instances, the two methods exhibit comparable performance.

%% file: tables/dataset.tex

%% file: tables/XNLI1.tex
\begin{table*}[t]
\centering
\caption{\textbf{Result comparison for cascading BERT, multilingual BERT base, XLM-RoBERTa base, and XLM-RoBERTa large on XNLI.} The blue rows represents a speed-up ration of 2. The red rows represents a speed-up ratio of 3. And the yellow rows represents a speed-up ration of 4.} 
\vskip -0.5em
\resizebox{2\columnwidth}{!}{%
\begin{tabular}{lccccccccccccccccc}
\toprule
  Method &
  ar  &
  bg &
  de &
  el &
  en &
  es &
  fr &
  hi &
  ru &
  sw &
  th &
  tr &
  ur &
  vi &
  zh &
  Avg. \\ \midrule
\rowcolor{LightCyan}
  PABEE &
  60.03 &
  65.23 &
  70.65 &
  82.97 &
  82.97 &
  76.44 &
  75.47 &
  69.06 &
  74.75 &
  57.84 &
  68.12 &
  71.11 &
  62.49 &
  73.87 &
  70.62 &
  69.92\\

\rowcolor{LightCyan}
  DeeBERT &
  38.32 &
  61.47 &
  46.13 &
  59.70 &
  58.90 &
  55.73 &
  56.27 &
  51.08 &
  48.42 &
  51.52 &
  45.71 &
  57.21 &
  55.05 &
  49.48 &
  48.42 &
  52.23\\
\rowcolor{LightCyan}
  CascadeBERT &
  71.36 &
  75.12 &
  77.06 &
  72.62 &
  86.72 &
  80.36 &
  79.18 &
  63.65 &
  74.10 &
  56.89 &
  56.76 &
  68.57 &
  62.48 &
  70.35 &
  73.59 &
  71.25\\

\rowcolor{LightCyan}
  Cascades &
  75.51 &
  79.64 &
  79.64 &
  78.24 &
  86.72 &
  \textbf{82.11} &
  81.31 &
  72.01 &
  77.98 &
  63.31 &
  70.71 &
  74.25 &
  68.04 &
  77.30 &
  75.90 &
  76.18\\

\rowcolor{LightCyan}
  {\name} &
  \textbf{75.75} &
  \textbf{81.00} &
  \textbf{80.74} &
  \textbf{79.94} &
  \textbf{87.31} &
  81.76 &
  \textbf{81.50} &
  \textbf{73.81} &
  \textbf{78.30} &
  \textbf{66.99} &
  \textbf{74.09} &
  \textbf{75.47} &
  \textbf{68.86} &
  \textbf{77.68} &
  \textbf{75.99}&
  \textbf{77.28}\\ \midrule

\rowcolor{LightRed}
  PABEE &
  52.89 &
  52.89 &
  62.07 &
  61.25 &
  76.03 &
  70.09 &
  68.8 &
  57.86 &
  62.99 &
  46.70 &
  60.56 &
  62.98 &
  53.21 &
  65.09 &
  63.89 &
  61.15\\
\rowcolor{LightRed}

  DeeBERT &
  39.90 &
  40.48 &
  40.58 &
  40.02 &
  49.50 &
  48.56 &
  49.20 &
  44.91 &
  42.47 &
  39.50 &
  41.86 &
  41.12 &
  39.48 &
  41.60 &
  45.19 &
  42.96\\
\rowcolor{LightRed}
  CascadeBERT &
  68.61 &
  71.26 &
  70.52 &
  68.51 &
  84.66 &
  76.99 &
  75.69 &
  57.58 &
  69.48 &
  51.66 &
  46.97 &
  63.07 &
  58.44 &
  64.69 &
  70.36 &
  66.56\\
\rowcolor{LightRed}

  Cascades &
  71.32 &
  76.39 &
  77.22 &
  74.39 &
  \textbf{86.48} &
  \textbf{80.19} &
  79.02 &
  66.78 &
  74.91 &
  59.24 &
  59.50 &
  69.70 &
  \textbf{64.43} &
  74.49 &
  \textbf{73.59} &
  72.51\\
\rowcolor{LightRed}

  {\name} &
  \textbf{71.44} &
  \textbf{78.44} &
  \textbf{78.60} &
  \textbf{75.87} &
  86.07 &
  79.68 &
  \textbf{79.46} &
  \textbf{68.38} &
  \textbf{76.67} &
  \textbf{62.16} &
  \textbf{67.76} &
  \textbf{72.04} &
  63.35 &
  \textbf{75.99} &
  73.53 &
\textbf{73.96}\\  \midrule

\rowcolor{LightYellow}
  PABEE &
  48.62 &
  48.62 &
  54.95 &
  53.58 &
  73.45 &
  63.39 &
  63.43 &
  51.85 &
  58.08 &
  44.27 &
  54.59 &
  57.28 &
  48.74 &
  60.92 &
  59.08& 
  56.06\\

\rowcolor{LightYellow}
  DeeBERT &
  38.92 &
  37.52 &
  39.76 &
  37.02 &
  43.39 &
  40.36 &
  41.04 &
  37.88 &
  40.12 &
  37.30 &
  38.48 &
  36.23 &
  35.91 &
  41.60 &
  39.58 &
  39\\
\rowcolor{LightYellow}
  CascadeBERT &
  63.49 &
  67.74 &
  69.40 &
  63.35 &
  83.64 &
  76.10 &
  74.86 &
  54.93 &
  68.89 &
  48.50 &
  42.43 &
  59.2 &
  55.63 &
  61.38 &
  68.44 &
  63.86\\

\rowcolor{LightYellow}
  Cascades &
  68.48 &
  73.11 &
  74.39 &
  70.58 &
  85.14 &
  \textbf{78.82} &
  77.26 &
  62.23 &
  72.14 &
  54.45 &
  54.59 &
  65.01 &
  \textbf{60.85} &
  70.23 &
  \textbf{71.75} &
  69.26\\

\rowcolor{LightYellow}
  {\name} &
  \textbf{69.88} &
  \textbf{75.07} &
  \textbf{75.51} &
  \textbf{72.42} &
  \textbf{85.45} &
  78.60 &
  \textbf{78.28} &
  \textbf{64.17} &
  \textbf{74.49} &
  \textbf{56.49} &
  \textbf{58.22} &
  \textbf{66.99} &
  60.82 &
  \textbf{72.30} &
  70.98 &
  \textbf{70.64}\\  \bottomrule
\end{tabular}%
}
\label{tab:xnli1}
\vspace{-0.5em}
\end{table*}

%% file: tables/PAWSX.tex
\begin{table*}[t]
\small
\caption{\textbf{Result comparison for cascading BERT, multilingual BERT base, XLM-RoBERTa base, and XLM-RoBERTa large on PAWS-X.} The blue rows represents a speed-up ration of 2. The red rows represents a speed-up ratio of 3. And the yellow rows represents a speed-up ration of 4.} 
\vskip -1em
\centering
\begin{tabular}{lccccccccccc}
\toprule
Method       & en    & fr   & es    & de    & zh     & ja    & ko   &  Avg. \\ \midrule
\rowcolor{LightCyan}
         PABEE        & 
         90.35&
         87.75&
         85.75&
         86.2&
         79.4&
         74.45&
         73.90& 
         82.54\\
\rowcolor{LightCyan}
         
  DeeBERT   &
  75.9&
  71.7&
  70.6&
  64.7&      
  64.8&      
  59.2&
  62.4&
  67.04\\
\rowcolor{LightCyan}
  CascadeBERT &
  92.25&
  88.3&
  87.7&
  87.8&
  79.75&
  75.6&
  75.88&
  83.89\\
\rowcolor{LightCyan}
 
  Cascades &
  \textbf{95.6} &
  \textbf{91.45} &
  90.85 &
  90.2 &
  84.4 &
  80.4 &
  79.85 &
  87.53\\
\rowcolor{LightCyan}
 
  {\name} &
  \textbf{95.6} &
  90.85 &
  \textbf{91.5} &
  \textbf{90.45} &
  \textbf{84.5} &
  \textbf{81.7} &
  \textbf{81.6} &
  \textbf{88.03}\\ \midrule
\rowcolor{LightRed}
         PABEE     &  63.55	& 63.3	& 57.6	& 58.25	& 58.35	& 62.25	& 59.2  & 60.36 \\
\rowcolor{LightRed}
         DeeBERT     &67.45	& 64.8	& 63.9	& 60.75	& 60.4	& 56.8	& 57.4 & 61.64\\
\rowcolor{LightRed}

  CascadeBERT &
  94.2&
  87.35&
  87.85&
  85.7&
  77.85&
  75.15&
  72.6&
  82.96\\
\rowcolor{LightRed}
         Cascades    & 95.2 & \textbf{90.8} & 90.2 & 90.05 & 81.65 & 77.7 & 76.75 & 86.05\\

\rowcolor{LightRed}
  {\name} &
  \textbf{95.6} &
  89.95 &
  \textbf{90.85} &
  \textbf{90.25} &
  \textbf{82.45} &
  \textbf{78.55} &
  \textbf{78.75} &
  \textbf{86.63} \\ \midrule
\rowcolor{LightYellow}
         PABEE        &  60.3	&63.2	& 57.25	& 57.4	& 57.75	& 56.8	& 58.1  & 58.69 \\
\rowcolor{LightYellow}
         DeeBERT     & 60	& 61.25	& 60.15 & 61& 58.15	& 56.1	& 56.65 & 59.04 \\
\rowcolor{LightYellow}
  CascadeBERT &
  94.25&
  85.55&
  85.7&
  84.05&
  75.9&
  73.1&
  70.5&
  81.29\\
\rowcolor{LightYellow}
         Cascades    & 95.2 & \textbf{89.75} & 89.3 & 88.3  & 80.3  & 75.7 & 74.75 & 84.76\\

\rowcolor{LightYellow}
  {\name} &
  \textbf{95.5} &
  88.8 &
  \textbf{89.45} &
  \textbf{88.45} &
  \textbf{80.55} &
  \textbf{77.35} &
  \textbf{76.55} &
  \textbf{85.24}\\ \bottomrule
\end{tabular}%
  \vspace{-1.5em}

\label{tab:pawsx}
\end{table*}

%% file: tables/QAM.tex
\begin{table}[t]
\small
\centering
\caption{\textbf{Result comparison for cascading BERT, multilingual BERT base, XLM-RoBERTa base, and XLM-RoBERTa large on QAM.} The blue rows represents a speed-up ration of 2. The red rows represents a speed-up ratio of 3. And the yellow rows represents a speed-up ration of 4.} 
\vskip -1em
\resizebox{0.9\columnwidth}{!}{%
\begin{tabular}{lcccc}
\toprule
 Method    & en    & de   & fr       & Avg.           \\ \midrule
\rowcolor{LightCyan}
         PABEE       & 64.3 & 64.12	& 65.75    &64.72           \\
\rowcolor{LightCyan}
         DeeBERT     & 63.83& 56.82	& 63.58    &61.41           \\
\rowcolor{LightCyan}
 CascadeBERT & 65.56          & 64.7           &              64.22     & 64.83               \\
\rowcolor{LightCyan}
         Cascades    & 71.41          & 68.4           & 68.83  & 69.55          \\
\rowcolor{LightCyan}
         {\name}          & \textbf{72.45} & \textbf{70.7}  & \textbf{70.19}  & \textbf{71.11} \\ \midrule
\rowcolor{LightRed}
         PABEE       & 60.6& 59.54	& 61.64      &60.59          \\
\rowcolor{LightRed}
         DeeBERT     &   57.8	& 53.32	& 57.36    &56.16           \\
\rowcolor{LightRed}
 CascadeBERT & 66.28          & 62.98          &              64.18        & 64.48               \\
\rowcolor{LightRed}
         Cascades    & 69.9           & 66.59          & 67.35     & 67.95          \\
\rowcolor{LightRed}
         {\name}          & \textbf{70.95} & \textbf{67.54} & \textbf{68.55} & \textbf{69.01} \\ \midrule
\rowcolor{LightYellow}
         PABEE        &  59.41	& 57.94	& 58.5  & 58.62            \\
\rowcolor{LightYellow}
         DeeBERT     &    56.38	& 51.99	& 55.67 &  54.68           \\
\rowcolor{LightYellow}
 CascadeBERT & 66.13          & 61.86          &              62.94         & 63.64              \\
\rowcolor{LightYellow}
         Cascades    & 68.52          & \textbf{64.64} & 65.73            & 66.3           \\
\rowcolor{LightYellow}
         {\name}          & \textbf{69.36} & 64.45          & \textbf{66.3}  & \textbf{66.7}  \\ \bottomrule
\end{tabular}%
}
\label{tab:qam}
\vspace{-1em}
\end{table}

%% file: tables/decoder_result.tex
\begin{table*}[t]
\centering
\small
\caption{\textbf{Zero-shot esult comparison for Llama2 on PAWS-X (Classification), GSM8k (Generation), and TabMWP (Generation)}. Cascade method accuracy around 2, 3, 4 times speed-up on PAWS-X cannot be measured in our experiment because the concentrated confidence scores of Cascade,
predominantly clustering around the value 1, resulting from the overconfident predictions by Llama-2. For example,
the Cascade model is bounded at the ceiling at a 4.89 speed-up ratio. Any ratio smaller than that number cannot be
measured as we alter the threshold.} 
\begin{tabular}{lcccccccccc}
\toprule 
Dataset & Speed-up & Model & en  & fr & es  & de   & zh   & ja  & ko & Avg.      \\ \toprule
\multirow{8}{*}{PAWS-X}
&
1x&
LLama-2-70B&
  78.2 &
  72.05 &
  71.15 &
  72.8 &
  67.6 &
  65.9 &
  64.7 &
  70.34
   \\ 
   \cmidrule{2-11}
   &
   \multirow{2}{*}{$\sim$2x}&
Cascade & 
  - &
  - &
  - &
  - &
  -&
  - &
  - &
  -
  
  \\
  &
  &{\name} 
&
  77.35 &
  69.45 &
  67.85 &
  65.95 &
  65.05 &
  61.05 &
  62.45 &
  67.02
  \\
   \cmidrule{2-11}
  &\multirow{2}{*}{$\sim$3x}&
Cascade &
  - &
  - &
  - &
  - &
  -&
  - &
  - &
  -
  \\ 
  &&{\name} &
  70.05 &
  68.05 &
  66.45 &
  63.9 &
  62.7 &
  59.85 &
  60.05 &
  64.44

  \\ 
   \cmidrule{2-11}
  &\multirow{2}{*}{$\sim$4x}&Cascade &
  - &
  - &
  - &
  - &
  -&
  - &
  - &
  -
\\
  && {\name} 
&
  67 &
  67.15 &
  65.1 &
  62.95 &
  61.3 &
  59.35 &
  59.6 &
  63.21
   \\ \midrule

   \multirow{8}{*}{GSM8k}&1x&
   LLama-2-70B&
  54.4 &
  - &
  42.6 &
  38.2 &
  26.6 &
  23.4 &
  19.6 &
  34.13
   \\ 
   \cmidrule{2-11}
   &\multirow{2}{*}{$\sim$2x}&Cascade & 
  30.4 &
  - &
  21.8 &
  18.2 &
  
  \textbf{15.4} &
  11.8&
  \textbf{11.8} &
  18.23

  \\
  &&{\name} &
  \textbf{33.6} &
  - &
  \textbf{26.4} &
  \textbf{22} &
  
  13.4 &
  \textbf{14.6} &
  10.4 &
  \textbf{20.06}
  \\
   \cmidrule{2-11}
  &\multirow{2}{*}{$\sim$3x}&Cascade & 
  22.4 &
  - &
  15.8 &
  11.4 &
  \textbf{12.2} &
  8.4 &
  8.4 &
  13.1

    \\ 
    &
   &{\name} &
  \textbf{25.4} &
  - &
  \textbf{20.8} &
  \textbf{15.2} &
  
  11.4 &
  \textbf{14.6}&
  \textbf{8.6}  &
  \textbf{16}
  \\ 
   \cmidrule{2-11}
  &\multirow{2}{*}{$\sim$4x}&Cascade &  
  18.8 &
  - &
  13.4 &
  10.2 &
  
  \textbf{10.4} &
  7.6&
  7&
  11.23
  \\
  &&{\name} &

  \textbf{22.2} &
  - &
  \textbf{16.8}&
  \textbf{12.8} &
  
  9.8 &
  \textbf{9} &
  \textbf{7.2} &
  \textbf{12.96}
   \\ \midrule
   \multirow{8}{*}{TabMWP}&1x&
   LLama-2-70B&
  65.2 &
  - &
  51.8 &
  39.8 &
  35.2 &
  38 &
  40.2 &
  45.03
   \\ 
   \cmidrule{2-11}
   &\multirow{2}{*}{$\sim$2x}&Cascade & 
  \textbf{64.8} &
  - &
  \textbf{46.4}&
  39.4 &
  
  29.8 &
  32.4 &
  30.8 &
  \textbf{40.6}

  \\
  &&{\name}  &
  57.8 &
  - &
  45.4 &
  \textbf{39.6} &
  \textbf{33.4} &
  \textbf{34.4} &
  \textbf{32.2} &
  40.47

  \\
   \cmidrule{2-11}
  &\multirow{2}{*}{$\sim$3x}&Cascade & 
  53.8 &
  - &
  44 &
  37.4 &
  28.8 &
  29.4 &
  26.4 &
  36.7

    \\ 
    &&{\name} &
  \textbf{55} &
  - &
  \textbf{44.4} &
  \textbf{38} &
  
  \textbf{31.8} &
  \textbf{31.8}&
  \textbf{29.2} &
  \textbf{38.3}

  \\ 
   \cmidrule{2-11}
  &\multirow{2}{*}{$\sim$4x}&Cascade &  
  \textbf{54}&
  - &
  42 &
  36.2 &
  
  27.8 &
  \textbf{30.6}&
  25.6&
  36.03
  \\
  &&{\name} &
  53.2 &
  - &
  \textbf{42.2}&
  \textbf{38.4} &
  
  \textbf{30.8} &
  \textbf{30.6} &
  \textbf{28} &
  \textbf{37.2}
   \\ \bottomrule
\end{tabular}%
\vspace{-1em}
\label{tab:llama}
\end{table*}

%% file: latex/sections/analysis.tex
\begin{figure*}[t]
    \begin{subfigure}[t]{\columnwidth}
      \centering
      \includegraphics[trim={0.2cm 0cm 0cm 1cm}, clip, width=0.5\linewidth]{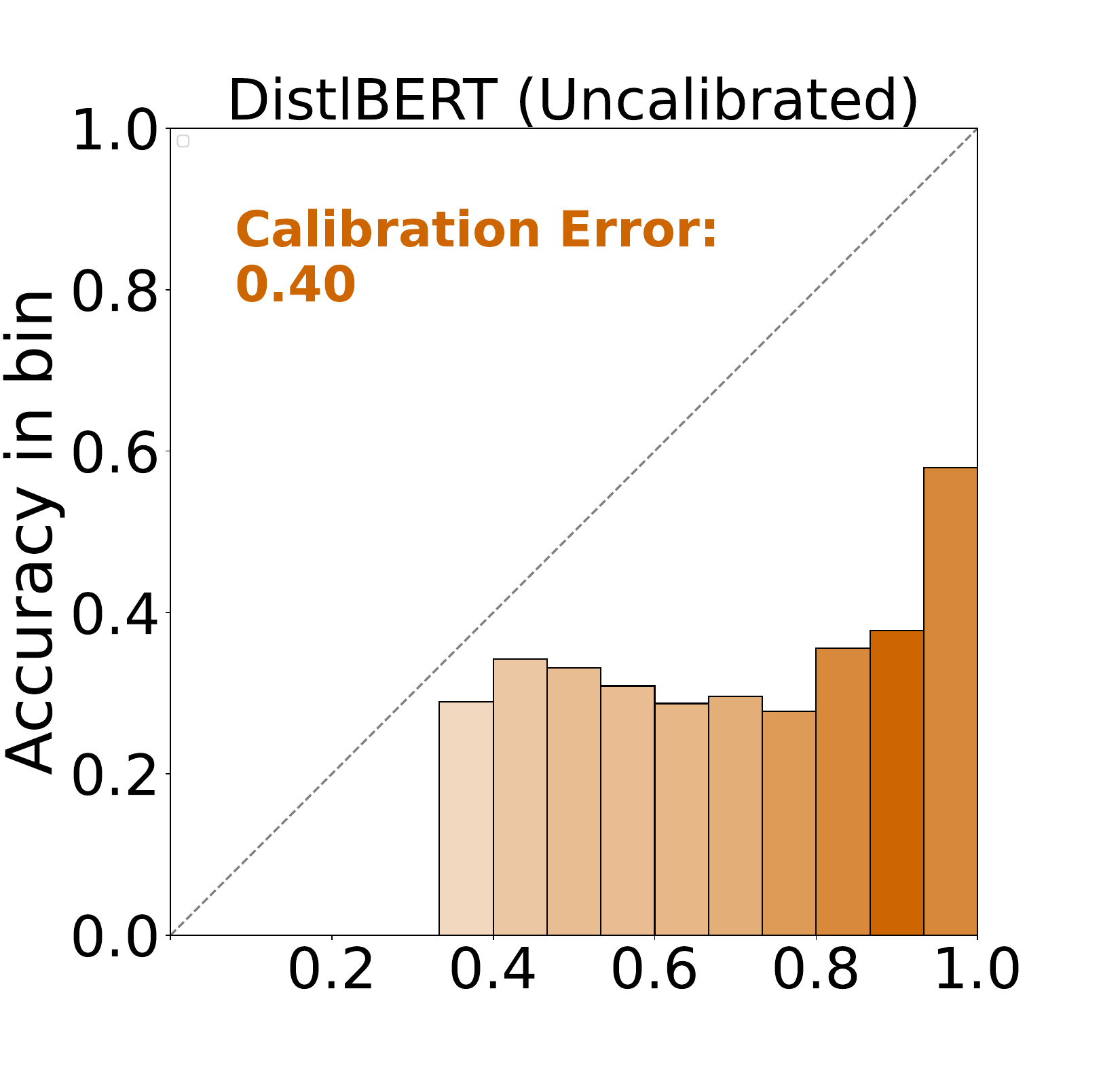}\hspace{-0.2cm}
      \includegraphics[trim={0.5cm 0cm 0cm 1cm}, clip, width=0.5\linewidth]{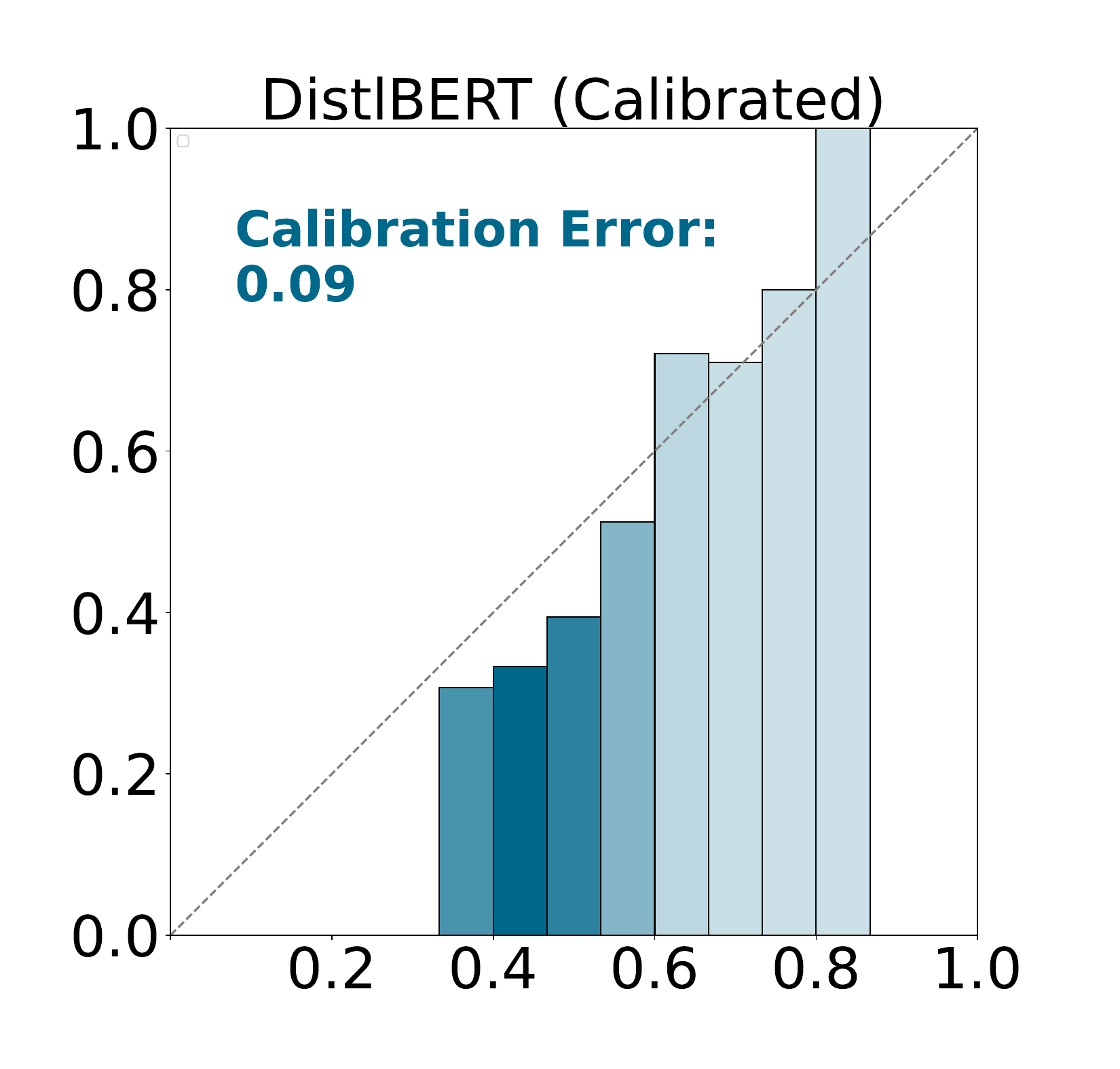}
      \includegraphics[trim={0.2cm 0cm 0cm 1cm}, clip, width=0.5\linewidth]{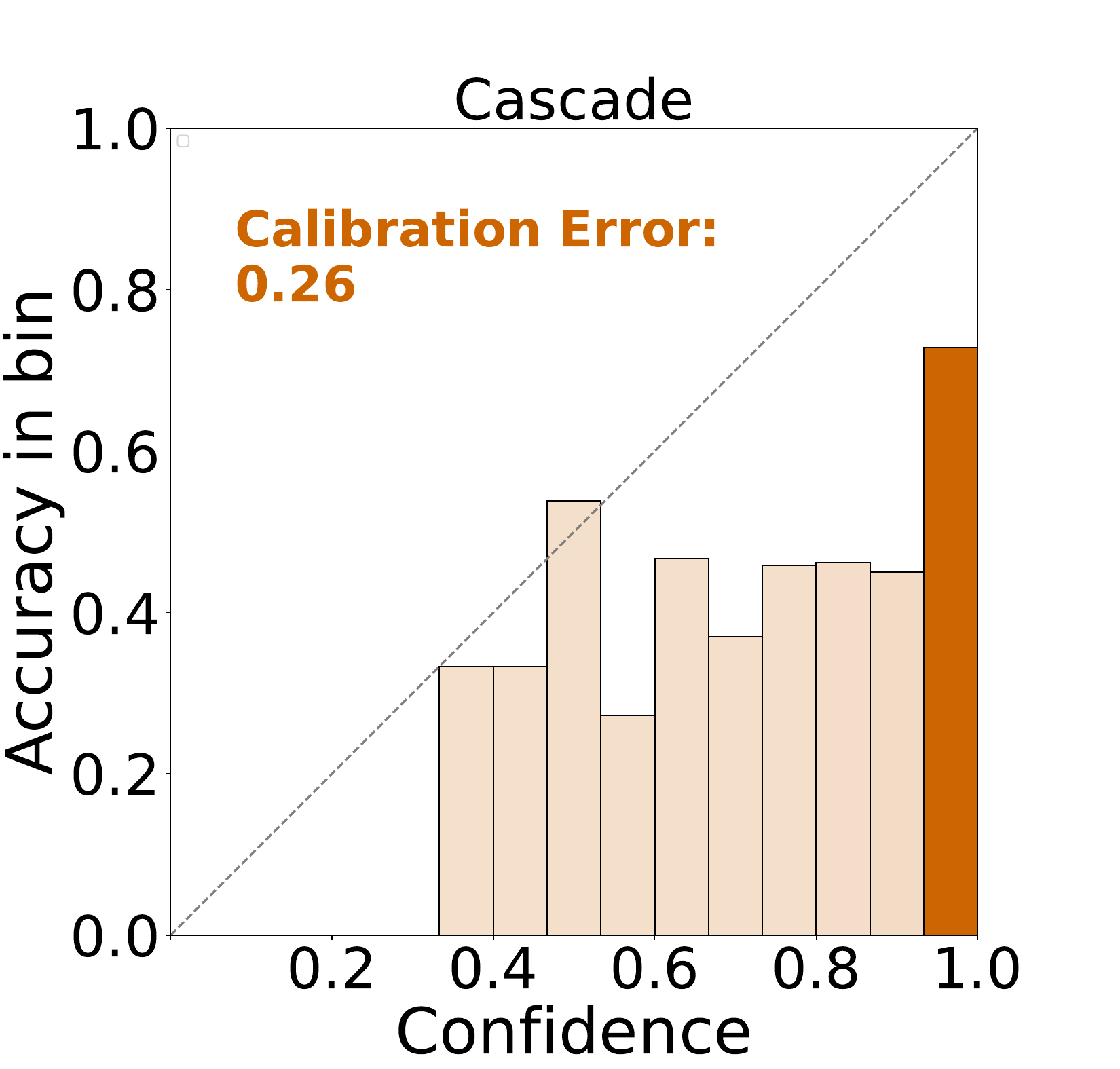} \hspace{-0.2cm}
      \includegraphics[trim={0.5cm 0cm 0cm 1cm}, clip, width=0.5\linewidth]{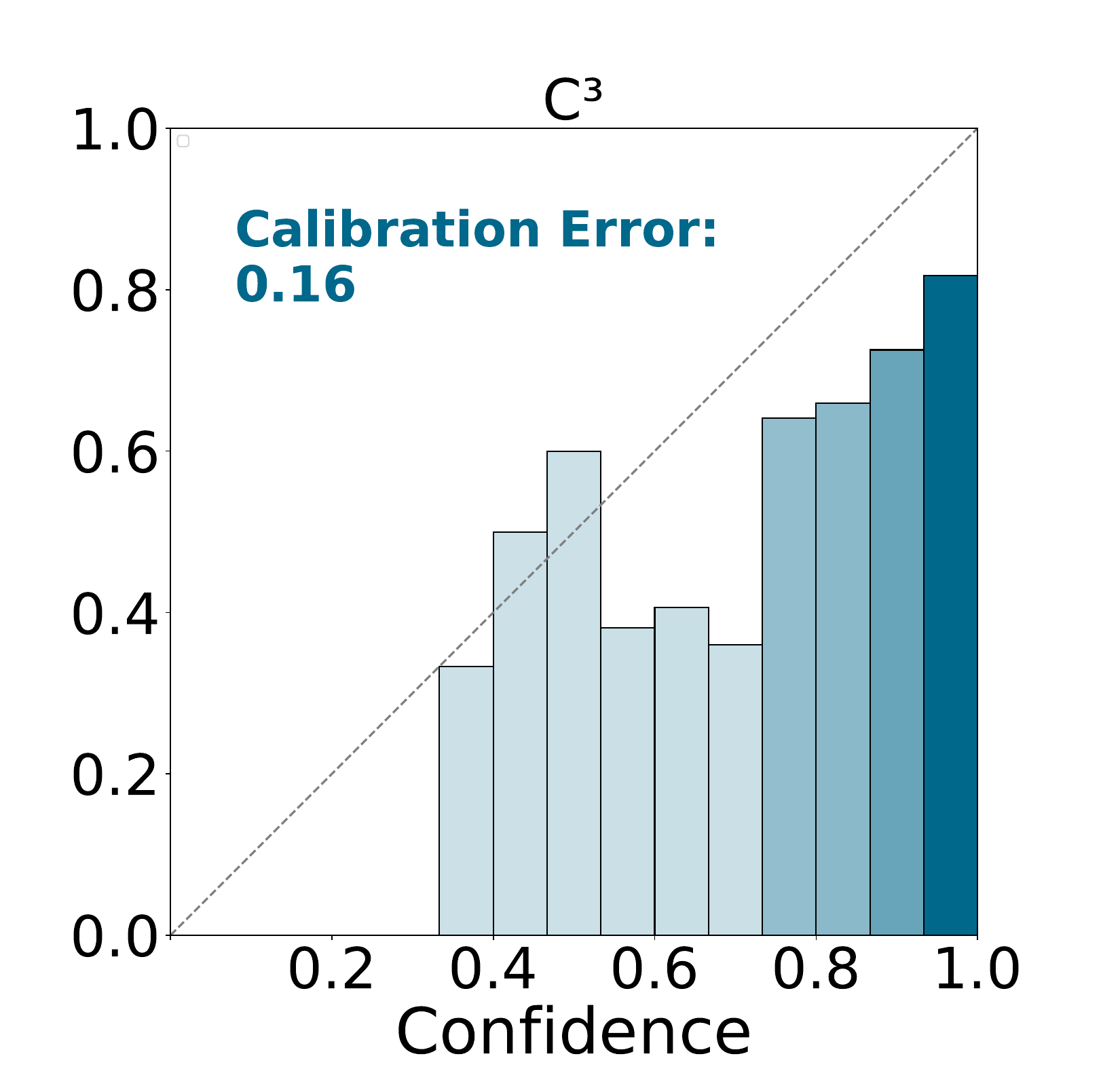} 
      \caption{Thai.}
      \label{fig:th_ece}
    \end{subfigure}
    \begin{subfigure}[t]{\columnwidth}
      \centering
      \includegraphics[trim={0.2cm 0cm 0cm 1cm}, clip, width=0.5\linewidth]{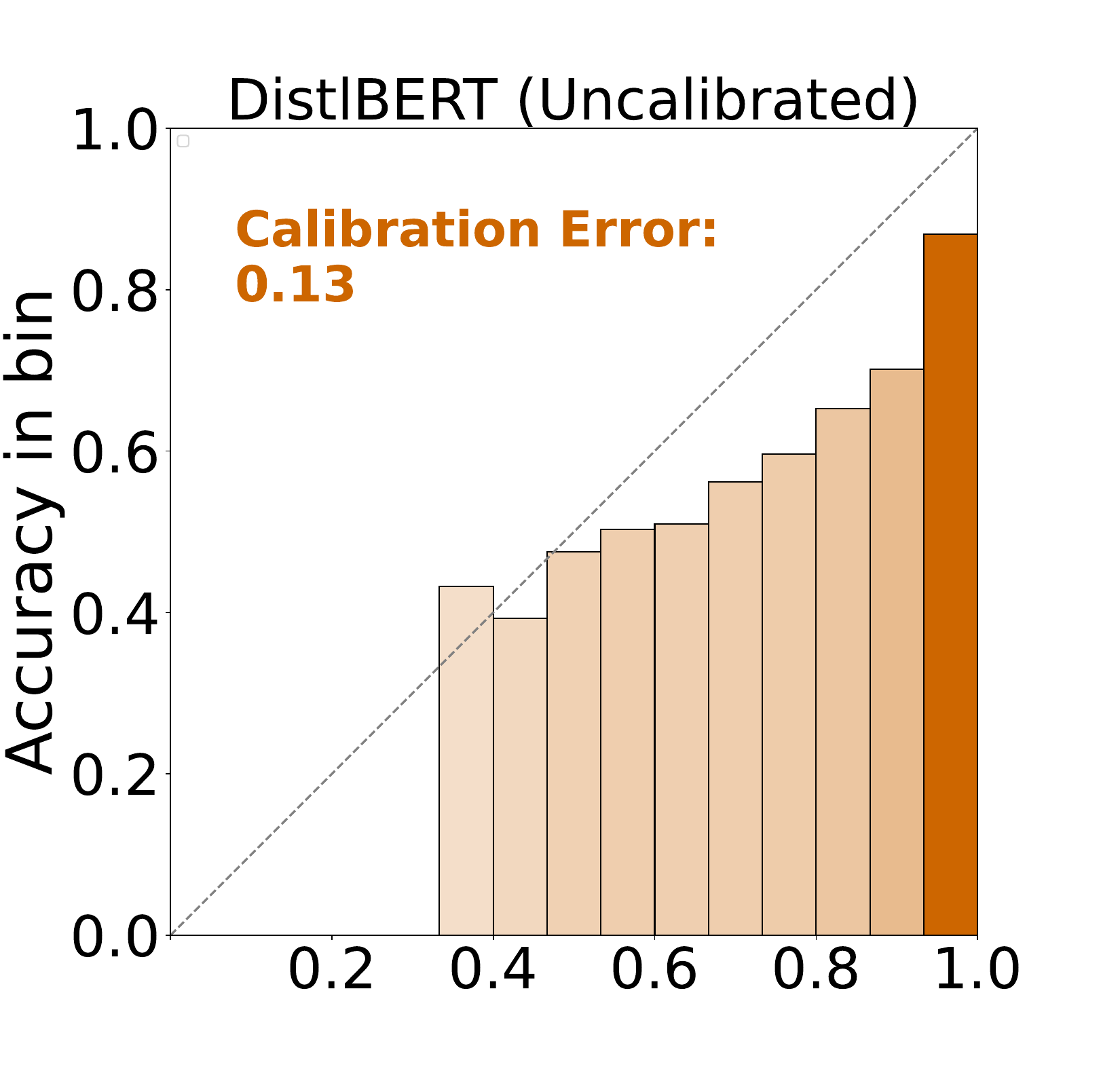}\hspace{-0.2cm}
      \includegraphics[trim={0.5cm 0cm 0cm 1cm}, clip, width=0.5\linewidth]{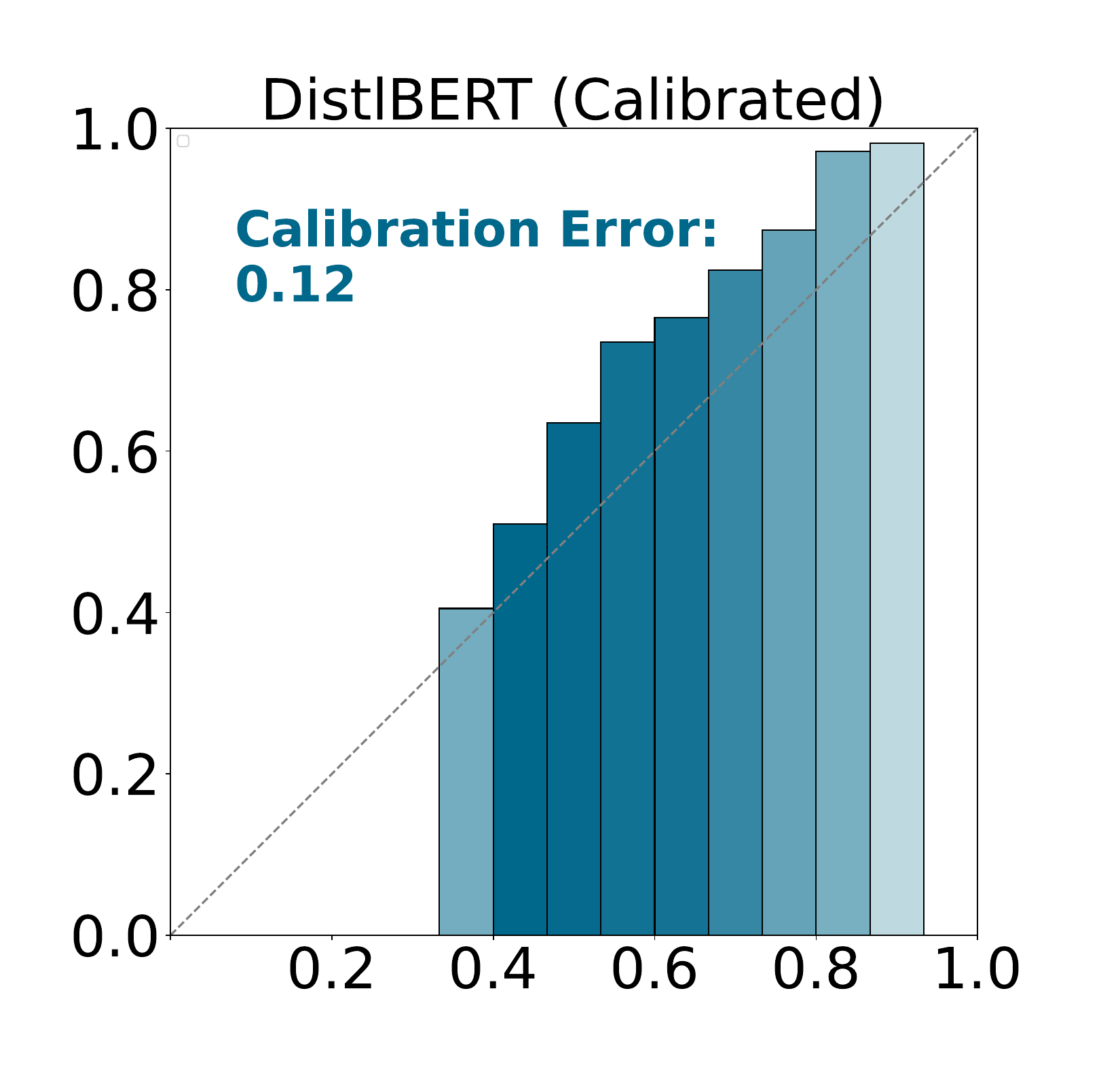}
      \includegraphics[trim={0.2cm 0cm 0cm 1cm}, clip, width=0.5\linewidth]{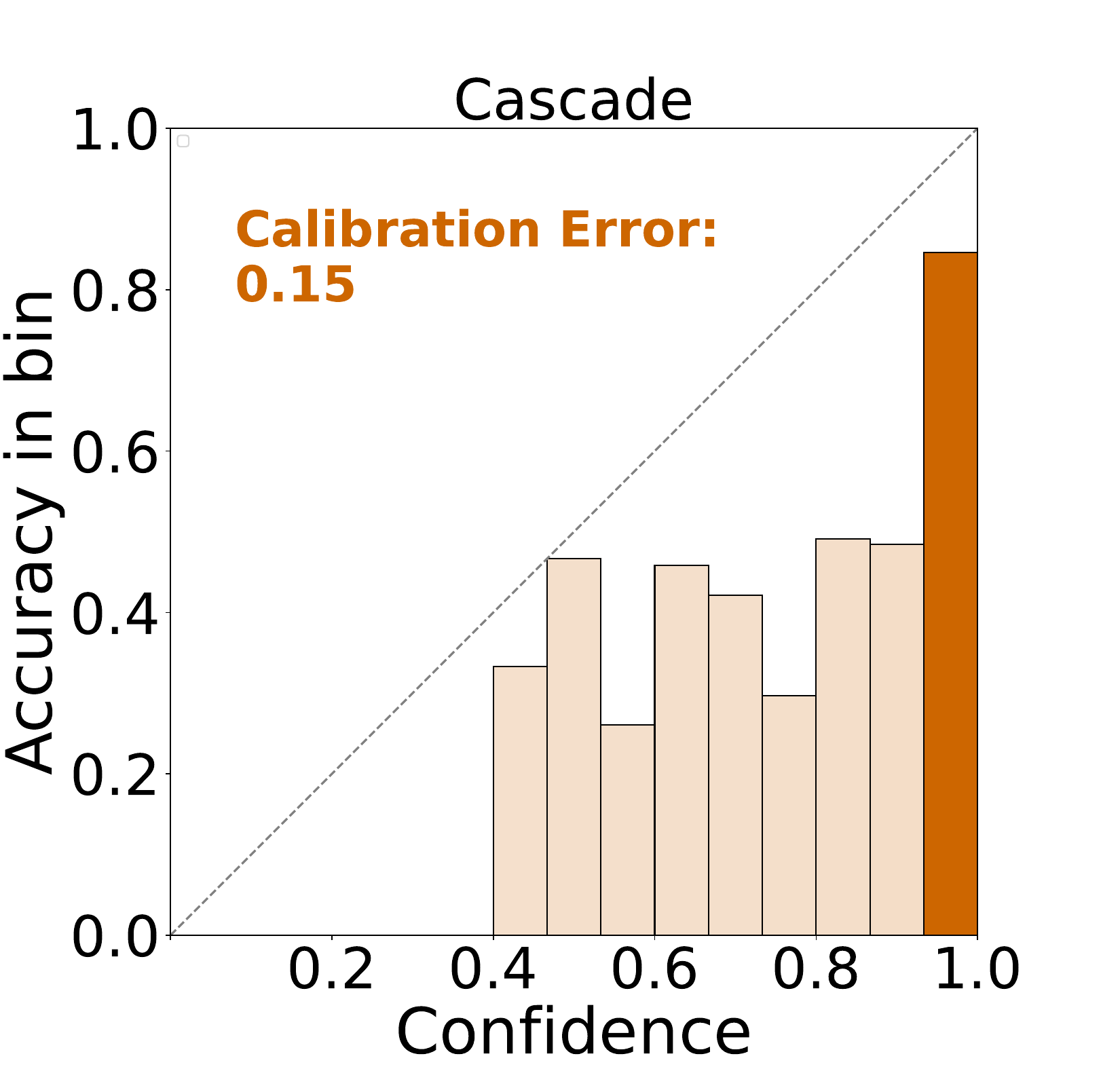} \hspace{-0.2cm}
      \includegraphics[trim={0.5cm 0cm 0cm 1cm}, clip, width=0.5\linewidth]{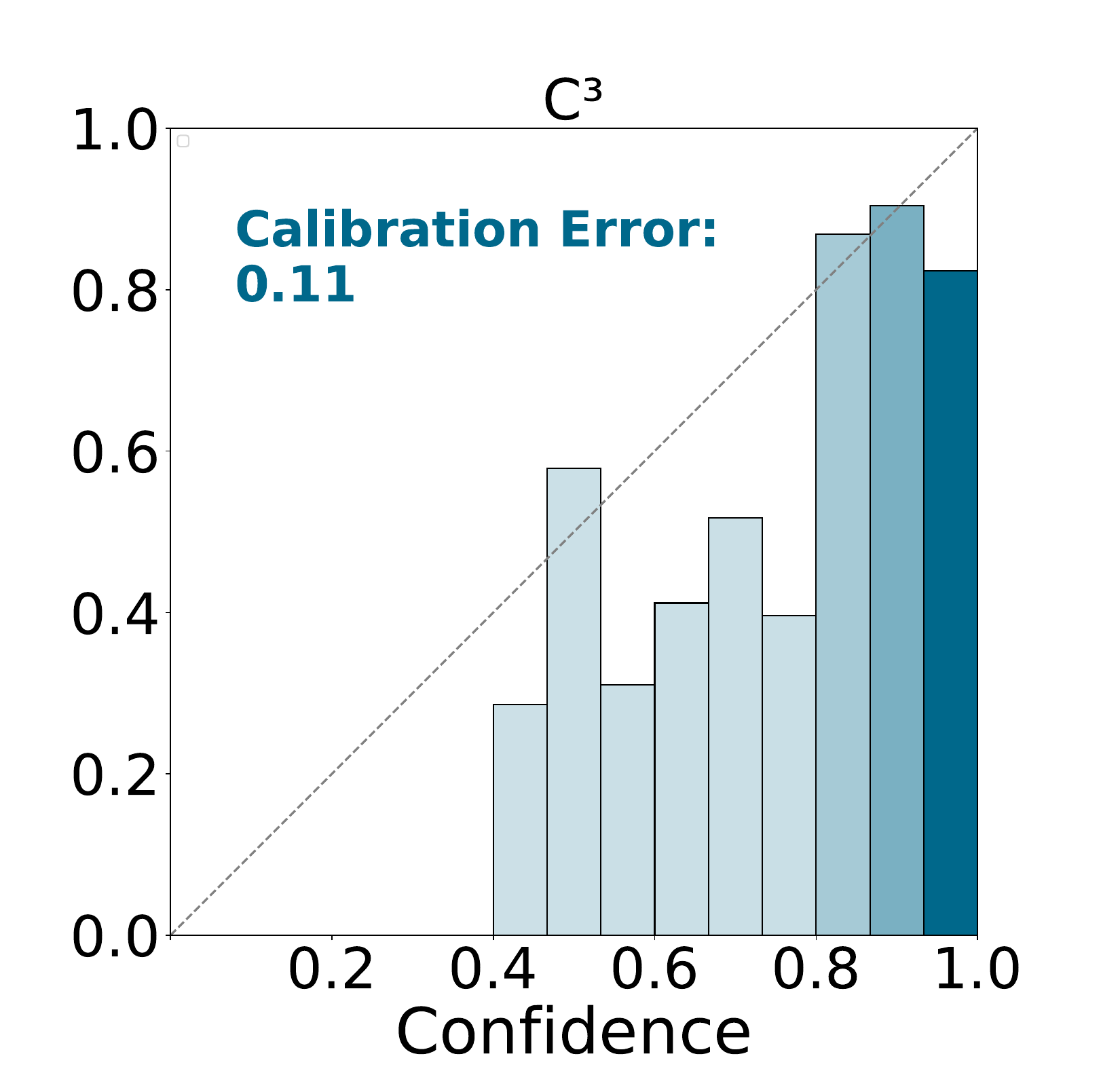} 
      \caption{Spanish.}
      \label{fig:es_ece}
    \end{subfigure}
    \vskip -1em
    \caption{\textbf{ECE comparison between Cascade (in orange) and {\name} (in blue).} In each subfigure, the top two figures are the ECE of the smallest model in these two methods, and the bottom two figures are the overall ECE.}
    \label{fig:ece}
    \vspace{-1em}
\end{figure*}

\subsection{Analysis}
 In this section, we study the effectiveness of the proposed calibration module to answer RQ2. We compute the calibration error of baseline Cascade and the proposed {\name} on XNLI with a 2x speedup ratio. We show the results on four languages, including Thai in Fig. \ref{fig:th_ece}, Spanish in Fig.  \ref{fig:es_ece}. See Appendix \ref{section:ece} for ECE of English and Swahili. Based on figures, we have following findings.

First, we observe that the proposed calibration method effectively reduces the calibration error for each individual model. In Fig.~\ref{fig:th_ece}, the expected calibration error (ECE) of the uncalibrated DistilBERT model on Thai data is notably high at approximately 0.4. Furthermore, we note a pattern of overconfidence in the uncalibrated model, with around 50\% of data having a confidence level exceeding 0.8, yet their actual accuracy is only around 40\%. This discrepancy is represented in the color depth of the histogram. In contrast, our proposed calibration method mitigates this overconfidence and miscalibration, as evidenced by the significantly reduced ECE of 0.09 for the calibrated DistilBERT, marking a substantial 78\% decrease. Calibration contributes to more reliable model confidence. A similar trend is observed in the case of Spanish data, as shown in Fig.~\ref{fig:es_ece}, where the ECE for the uncalibrated DistilBERT is 12.62, and after calibration, it decreases to 11.88.

Second, the entire model benefits from the calibration process. As depicted in Fig.~\ref{fig:th_ece}, the ECE for uncalibrated model cascade is 25.81, whereas the ECE for calibrated cascade is reduced to 15.75, representing a substantial decrease of approximately 40\% in ECE. Examining accuracy, the calibrated cascade achieves an accuracy of 74.09\%, which is 3.38\% higher than that of the uncalibrated cascade. Similar trends are observed in Spanish data, where the ECE for the uncalibrated cascade is 15.31, and the ECE for the calibrated cascade is 11.10.

%% file: latex/sections/hyper.tex
\subsection{Sensitivity w.r.t. Hyper-parameters}
 
\begin{figure}[!h]
    \centering
    \includegraphics[trim={0 0 0 2.7cm},clip,width=0.75\linewidth]{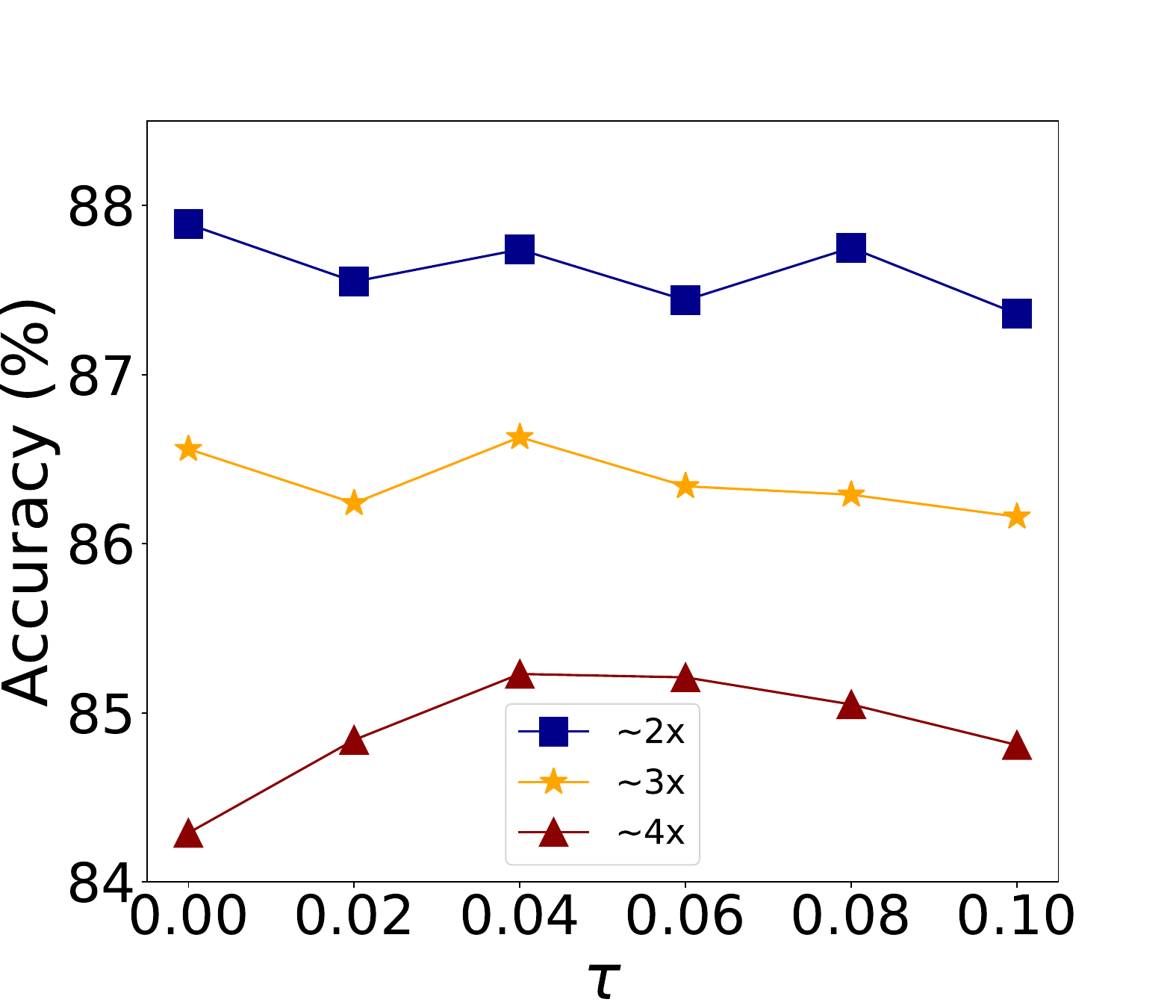}
    \vskip -0.5em
    \caption{{\name} accuracy on PAWS-X w.r.t. $\tau$ values.}
    \label{fig:tau}
    \vspace{-1em}
\end{figure}

We explore the sensitivity of our method concerning the parameter $\tau$ in logit normalization. In Fig.~\ref{fig:tau}, we present the average accuracy of our model on the PAWS-X dataset with varying values of $\tau$. With a two-fold speed-up, $\tau$ exerts minimal influence on cascade accuracy. However, when a higher speed-up ratio is required, we observe that a $\tau$ value of 0.04 corresponds to the peak accuracy.

\subsection{Case Study}

In this section, we present concrete cases to illustrate the functioning of the proposed {\name}. We examine several examples across different languages, including English, Bulgarian, German, Thai, and Chinese, where calibration rectifies predictions, as demonstrated in Figure \ref{fig:case} in the appendix. These cases reveal that vanilla cascade models frequently exhibit overconfidence, potentially leading to the selection of smaller models in error. In contrast, the proposed {\name} adeptly chooses appropriate models for inference. For instance, when presented with the premise "They said we're providing accommodations for your stay" and the hypothesis "They're covering housing costs," the cascade model displays unwavering confidence in the prediction from the second model, while {\name} selects the largest model. This highlights the prevalence of miscalibration, which often results in excessive overconfidence, causing the model to emit a prediction without progressing to larger models. Calibration plays a pivotal role in mitigating this issue, a pattern consistently observed across various languages.

%% file: latex/sections/conclusion.tex
\section{Conclusion}
In this paper, we propose a simple, yet effective method, \textbf{C}onfidence \textbf{C}alibration \textbf{C}ascade ({\name}) that enhances cross-lingual inference accuracy through more reliable confidence scores. We introduce calibration methods for both Language Model cascade and Large Language Model cascade. We conduct extensive experiment on cross-lingual benchmarks. By comparing with state-of-the-art methods, the results demonstrates the effectiveness of {\name}. Furthermore, {\name} also demonstrates strong calibration results compared to vanilla cascade methods.

%% file: latex/sections/limitation.tex
\section*{Limitations}
The proposed {\name} framework depends on a hyperparameter $\tau$ at a tuning process, which might require additional effort to tune. Fortunately, our experiment shows that the {\name} model is not sensitive to this hyperparameter, and thus this issue can be easily addressed.
\section*{Acknowledgement}
We thank Google Cloud Research Credits program for supporting our computing needs.

%% file: latex/sections/appendix.tex
\appendix
\section{Related Work}
\label{sec:appendix_related_work}
\noindent\textbf{Model Compression-based Methods.} There are a number of model compression methods which can be applied to accelerate PLM inference, such as knowledge distillation, pruning, and weight quantization. Knowledge distillation is to leverage a powerful teacher model to guide the learning of a lightweight student model. The lightweight student can support efficient inference. For example, DistilBERT~\cite{sanh2019distilbert}, TinyBERT~\cite{jiao-etal-2020-tinybert}, MobileBERT~\cite{sun-etal-2020-mobilebert}, and PKD~\cite{pkd} use knowledge distillation to learn a lightweight BERT. \cite{liang2023less}, \cite{zhou2023lima}, \cite{gu2023knowledge}, \cite{gu2023knowledge} and \cite{shridhar2023distilling} apply knowledge distillation to train a small generative model. Pruning focuses on finding redundant parameters and sets them as zero to achieve highly sparse neural network. For example, EBERT~\cite{liu2021ebert} and \cite{chen2020lottery} propose a dynamic structured pruning method and a lottery ticket hypothesis~\cite{frankle2018lottery} for BERT, and Sheared Llama~\cite{xia2023sheared} designs a structured pruning method for generative pre-trained language models. Weight quantization is to map the model weights into low-precision integers or floating-point numbers, which are hardware-friendly and efficient for matrix computation. Specifically, Q8BERT~\cite{zafrir2019q8bert} proposes 8-bit quantization for BERT; MKQ-BERT~\cite{tang2022mkq} designs 4-bit quantization method for BERT; BinaryBERT~\cite{bai2020binarybert} explore how to quantize BERT in 1 bit; Smoothquant~\cite{xiao2023smoothquant} and nuQmm~\cite{park2022nuqmm} study how to apply quantization for generative language models. However, these methods often need to train compressed models from scratch, which is expensive. Pruning and quantization rely on specialized hardware support, which is not flexible.
\begin{table}[t]
\begin{minipage}{1\columnwidth}  
    \begin{tcolorbox} 
        \hspace{-6mm}
        \begin{tabular}{p{0.99\columnwidth}}
        \hspace{1mm}
        \begin{minipage}{0.99\columnwidth}
        \textbf{User} \\
        Do (Sentence1, \textcolor{blue}{Example1: Sentence1}) and (Sentnece2, \textcolor{blue}{Example1: Sentence2}) have different meaning?

        Answer: \textcolor{blue}{Example1: Answer}. \\

        Do (Sentence1: \textcolor{blue}{Example2: Sentence1}) and (Sentence2: \textcolor{blue}{Example2: Sentence1}) have different meaning?

        Answer: \textcolor{blue}{Example2: Answer}. \\

        Do (Sentence1: \textcolor{blue}{Example2: Sentence1}) and (Sentence2: \textcolor{blue}{Example3: Sentence1}) have different meaning?

        Answer: \textcolor{blue}{Example3: Answer}. \\

        Do (Sentence1: \textcolor{red}{Question: Sentence1}) and (Sentence2: \textcolor{red}{Question: Sentence2})
         have different meaning?

        Answer:\\
        \rule[0.25\baselineskip]{\textwidth}{1pt}
        \textbf{Assistant} \\
        \textcolor{blue}{\texttt{Answer}}
        \end{minipage}
        \end{tabular}
    \end{tcolorbox}
    \vspace{-2mm}
    \caption{\textbf{{\name} prompt on Llama-2 for PAWS-X task.} The answer candidates set would be \{Yes, No\}, and the embedding set would be \{y, n\}, where y is the Llama embedding of the word "Yes," and n is the Llama embedding for the word "No."}
    \label{tab:prompt}
\end{minipage}
\end{table}

\section{Prompt Design}
\label{sec:appendix}
We include the prompt we use for Llama-2 in our experiment on PAWS-X in this section. Detailed prompt is shown in Table \ref{tab:prompt}.

\section{Algorithm}
\begin{algorithm}[h!]
\caption{Calibration Cascade Existing}
\label{alg:1}
\KwIn{Models $\{M_1,...,M_n\}$, threshold $\lambda$, data $x$}
\For {$i \gets 1$ \KwTo $n$}{
  // get the logits after temperature scaling\\
  logits = $M_i$($x$) / temperature($M_i$)\\
  // get the probability for each class label\\
  Pr = softmax(logits)\\
  // get the confidence\\
  confidence = max(Pr)\\
  \If{confidence > $\lambda$ or $i$ == $n$}{
    return logits
  }
}
\end{algorithm}

\begin{table}[h!]
\centering
\caption{Dataset we incorparate in our experiment.}
\resizebox{\columnwidth}{!}{
\begin{tabular}{ccccc}
\toprule
\textbf{\large{\multirow{2}{*}{Dataset}}}  & \textbf{\large{\# Train}}  &  \textbf{\large{\# Dev}}& \textbf{\large{\# Test}} & \textbf{\large{\multirow{2}{*}{Languages}}} \\
& (English)&(per langauge)&(per langauge)\\ 
\midrule
XNLI& 393k & 2.49k & 5.01k & 15 \\ 
\midrule
PAWS-X & 49.4k & 2k & 2k& 7 \\
\midrule
QAM & 100k & 10k & 10k & 3 \\ 
\midrule
GSM8k & \multirow{2}{*}{-}& \multirow{2}{*}{-} & \multirow{2}{*}{500} & \multirow{2}{*}{6}\\ 
(translated) & & & &\\
\midrule
TabMWP & \multirow{2}{*}{-}& \multirow{2}{*}{-} & \multirow{2}{*}{500} & \multirow{2}{*}{6}\\ 
(translated) & & & &\\
\bottomrule
\end{tabular}
}
\vspace{-1em}
\label{tab:dataset}
\end{table}
\begin{figure*}[h!]
    \centering
    \hfill
    \includegraphics[width=1.0\linewidth]{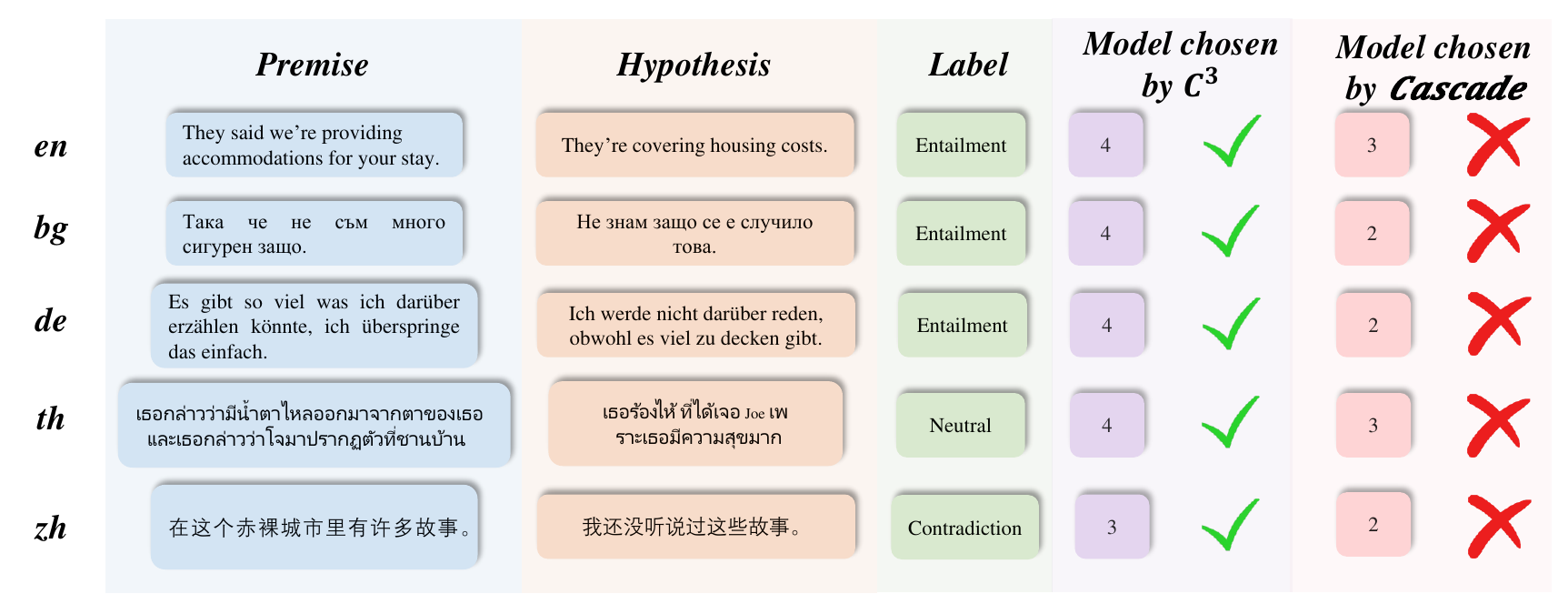}
    \vspace{-8mm}
    \caption{Case study of five languages.}
    \label{fig:case}
\end{figure*}

\section{ECE Figures}
\label{section:ece}
\begin{figure}[]
      \centering
      \includegraphics[trim={0.2cm 0cm 0cm 1cm}, clip, width=0.5\linewidth]{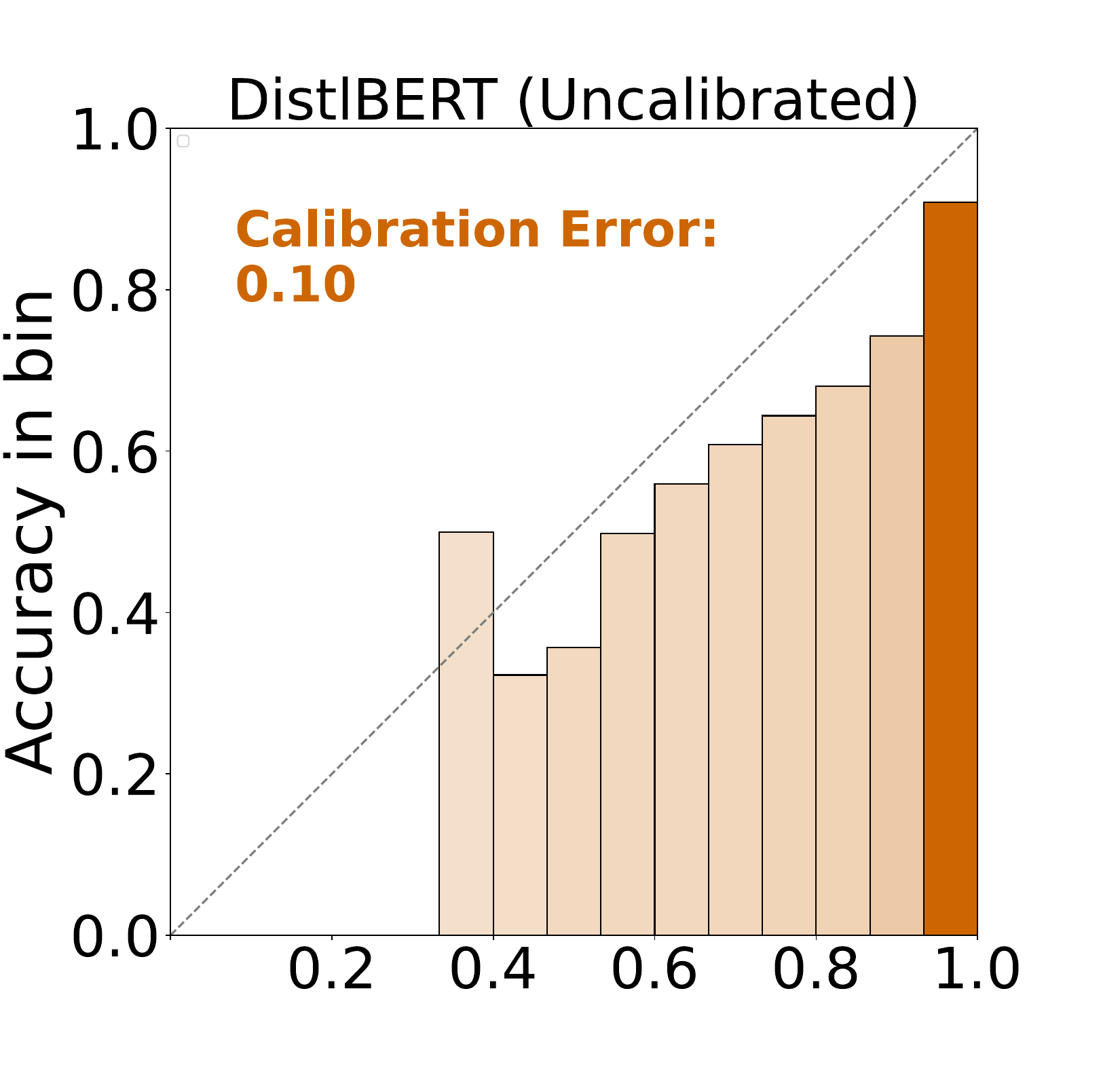}\hspace{-0.2cm}
      \includegraphics[trim={0.5cm 0cm 0cm 1cm}, clip, width=0.5\linewidth]{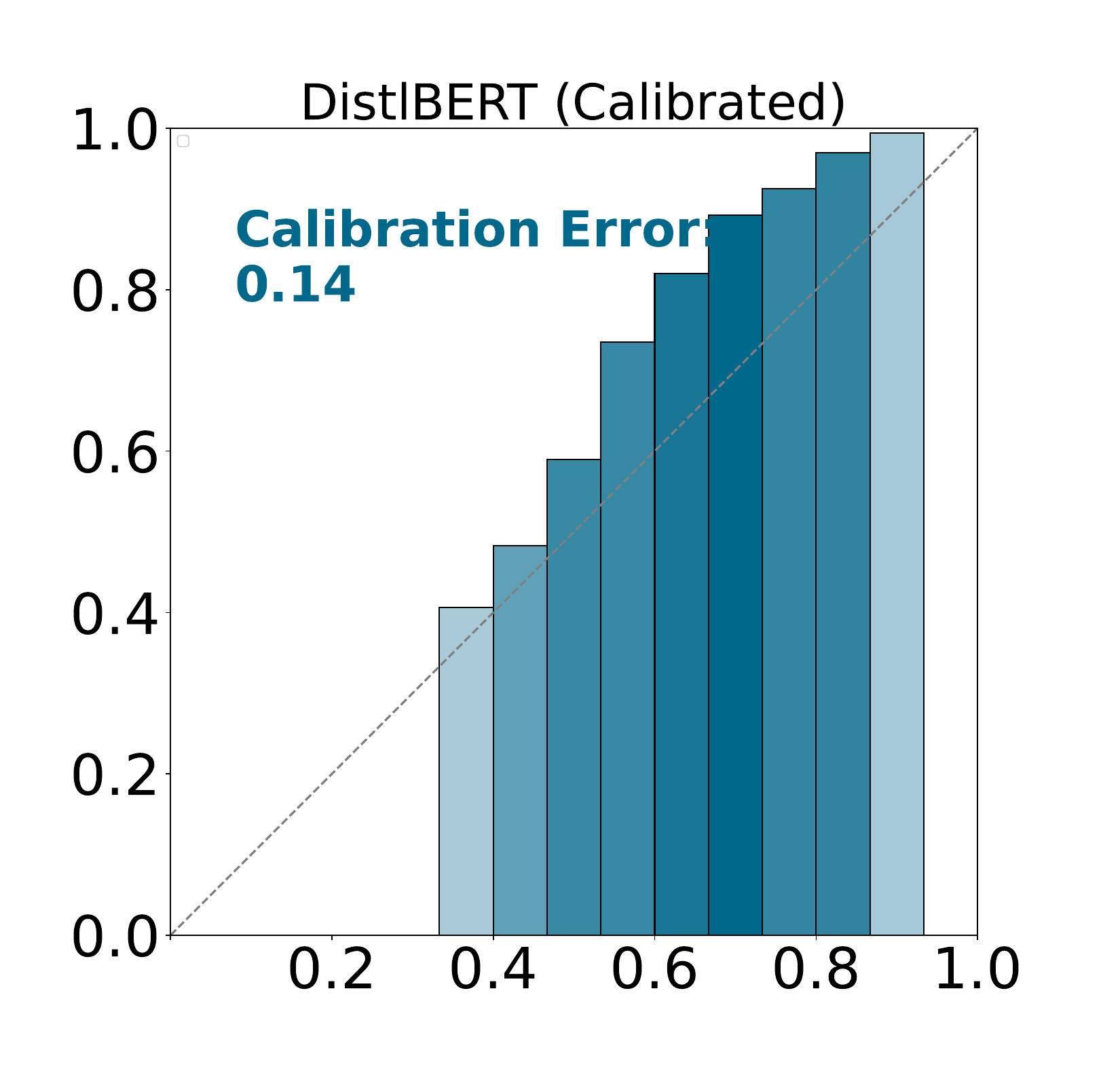}
      \includegraphics[trim={0.2cm 0cm 0cm 1cm}, clip, width=0.5\linewidth]{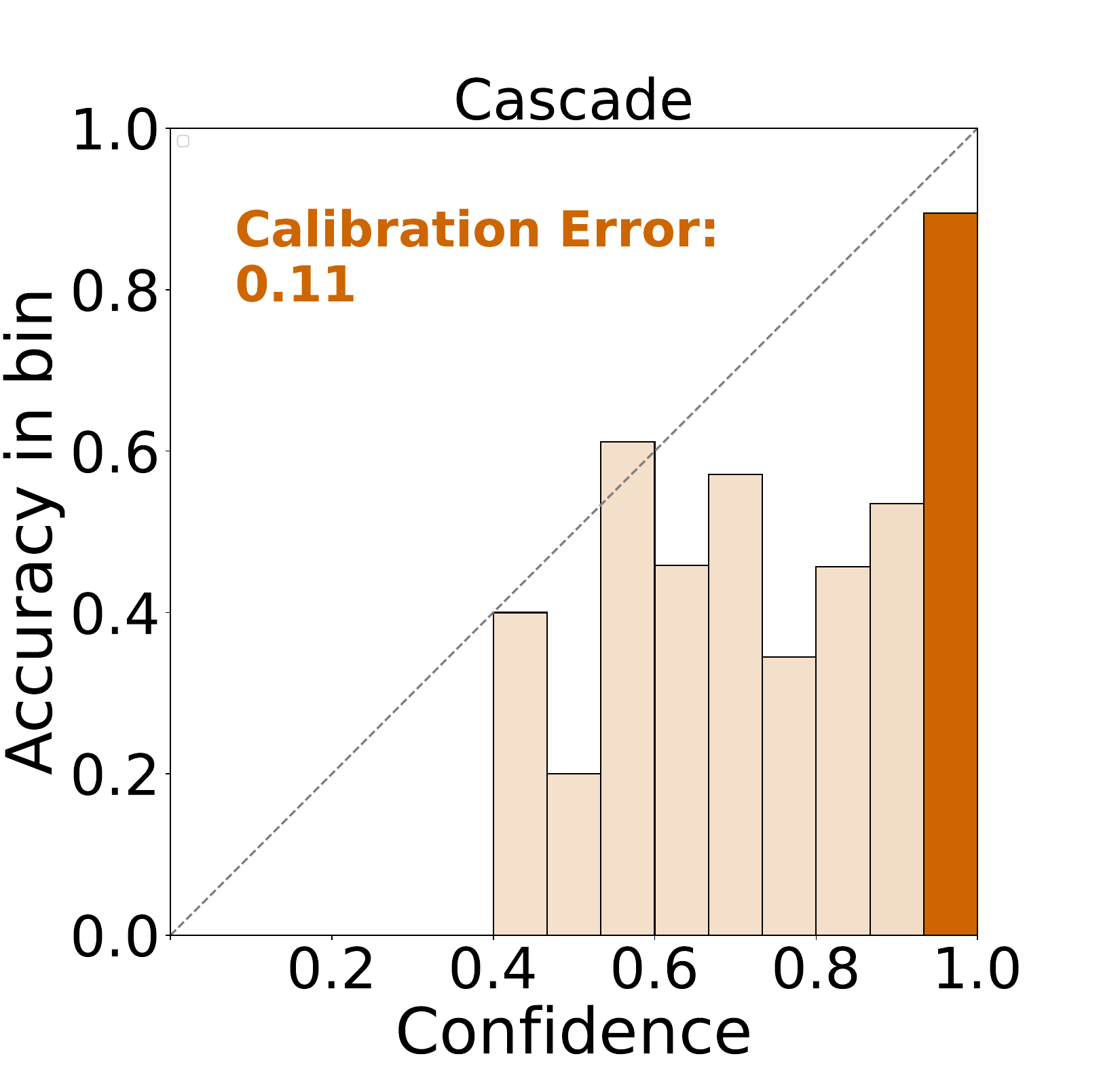} \hspace{-0.2cm}
      \includegraphics[trim={0.5cm 0cm 0cm 1cm}, clip, width=0.5\linewidth]{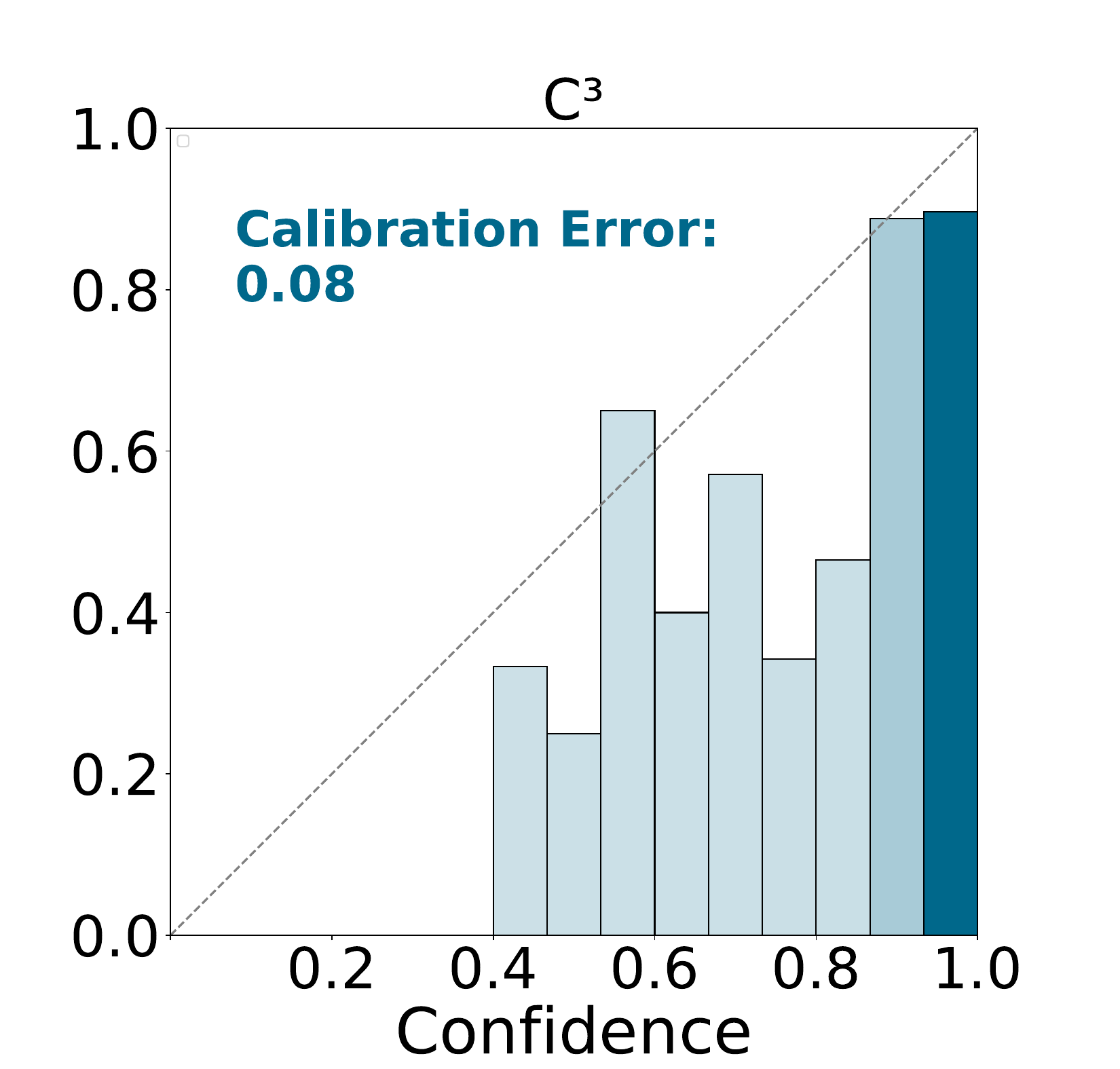}
      \caption{ECE comparison between Cascade and {\name} in English language.}
      \label{fig:en_ece}
\end{figure}

\begin{figure}[t]
      \centering
      \includegraphics[trim={0.2cm 0cm 0cm 1cm}, clip, width=0.5\linewidth]{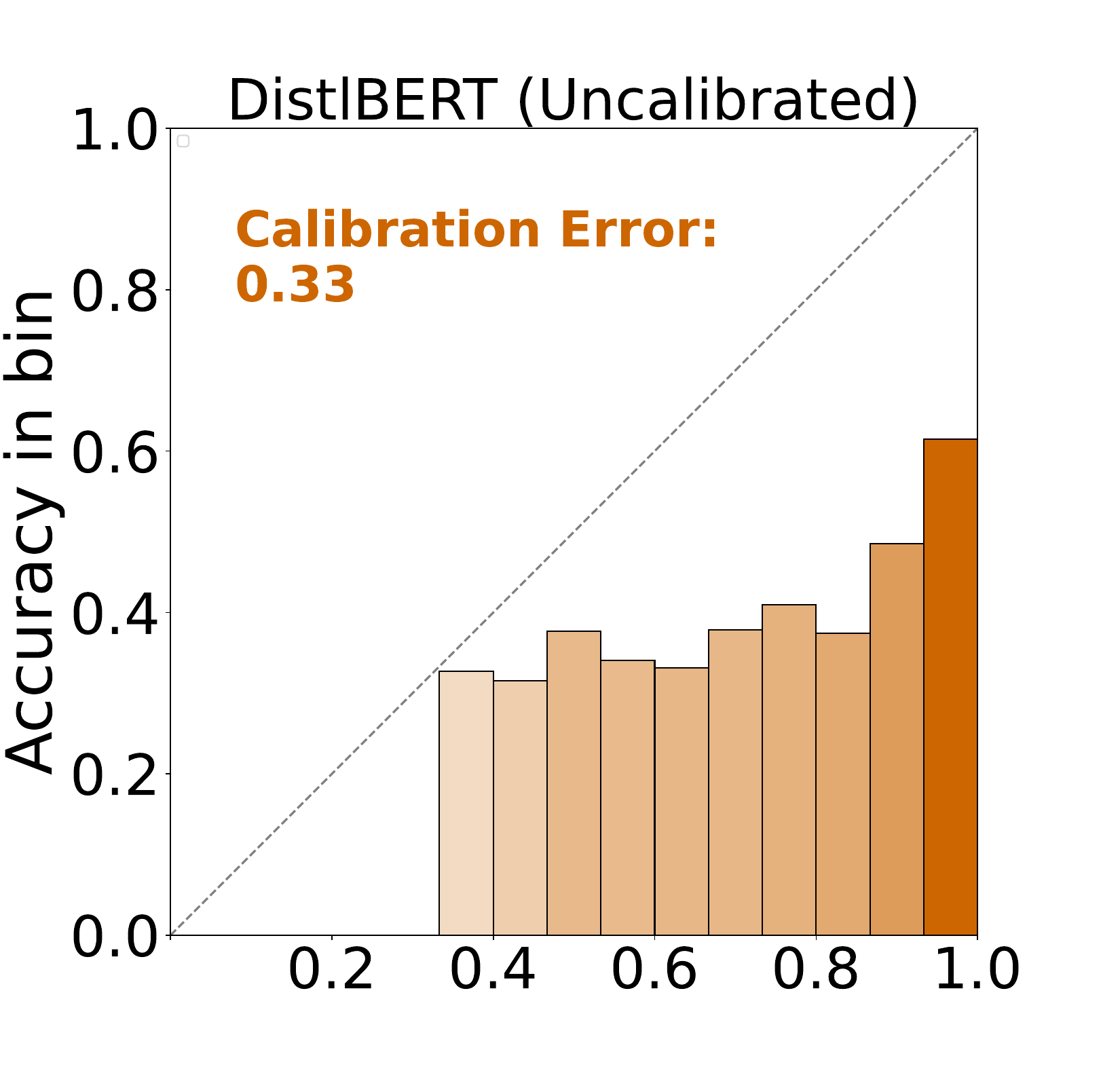}\hspace{-0.2cm}
      \includegraphics[trim={0.5cm 0cm 0cm 1cm}, clip, width=0.5\linewidth]{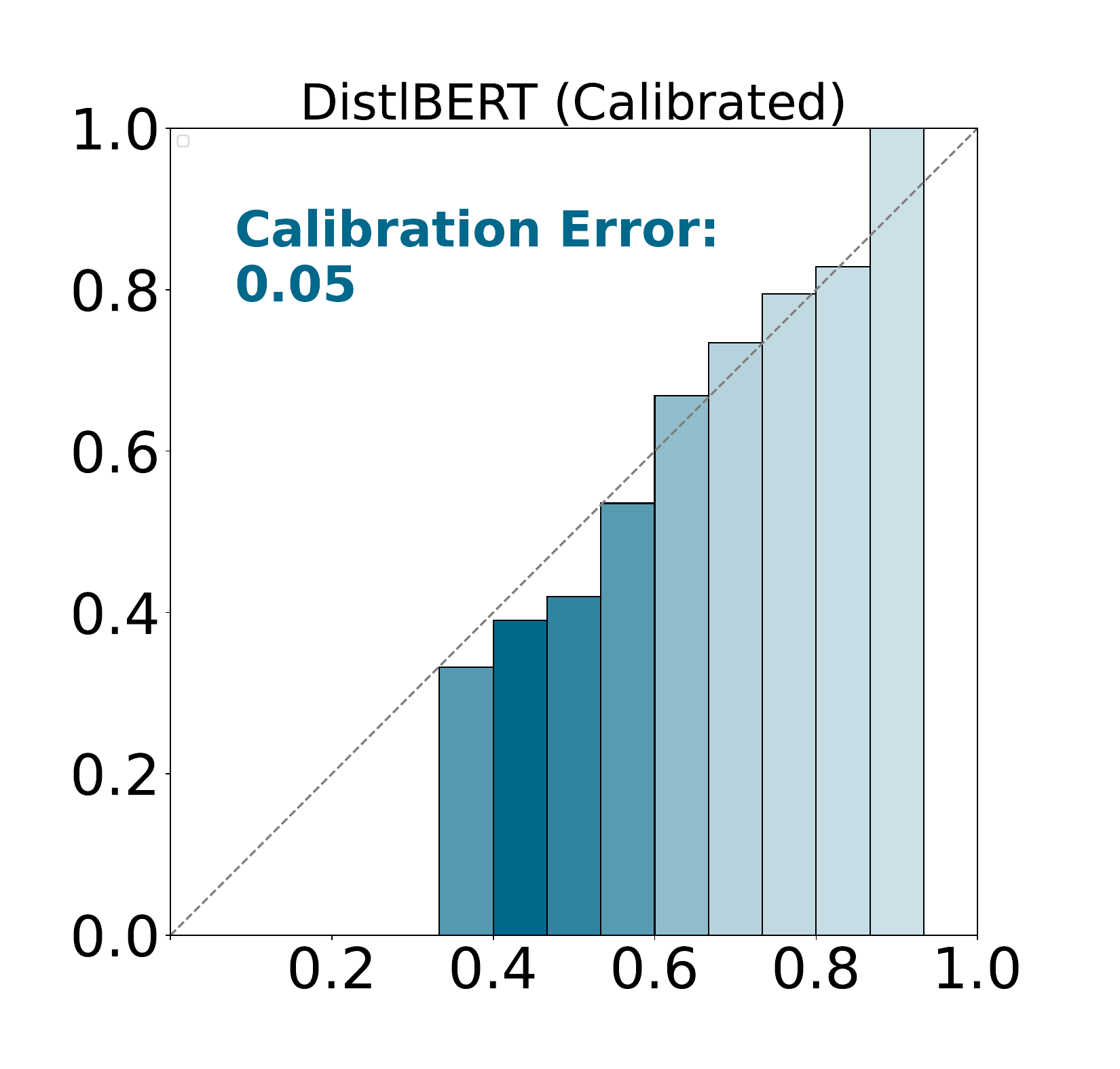}
      \includegraphics[trim={0.2cm 0cm 0cm 1cm}, clip, width=0.5\linewidth]{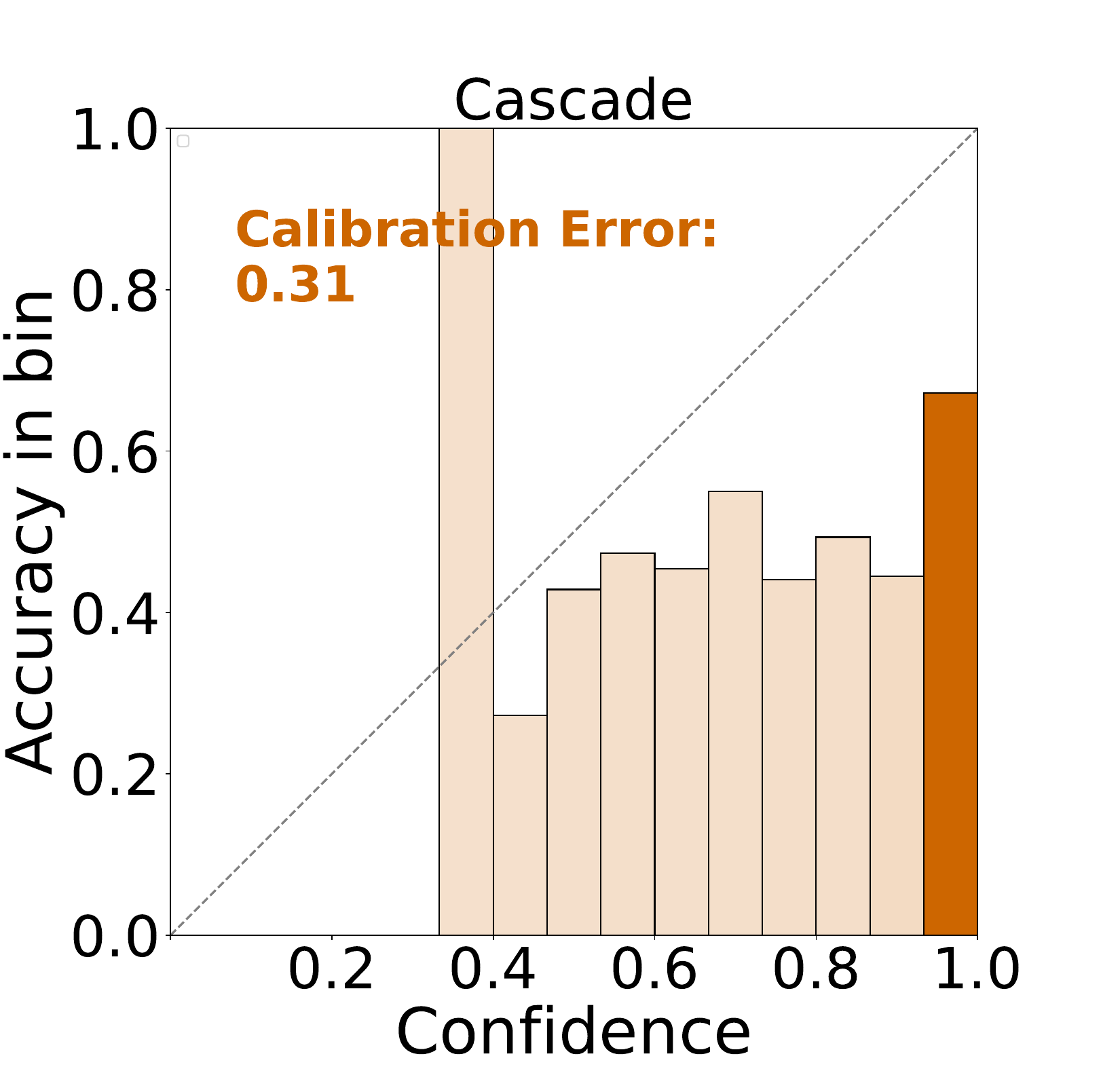} \hspace{-0.2cm}
      \includegraphics[trim={0.5cm 0cm 0cm 1cm}, clip, width=0.5\linewidth]{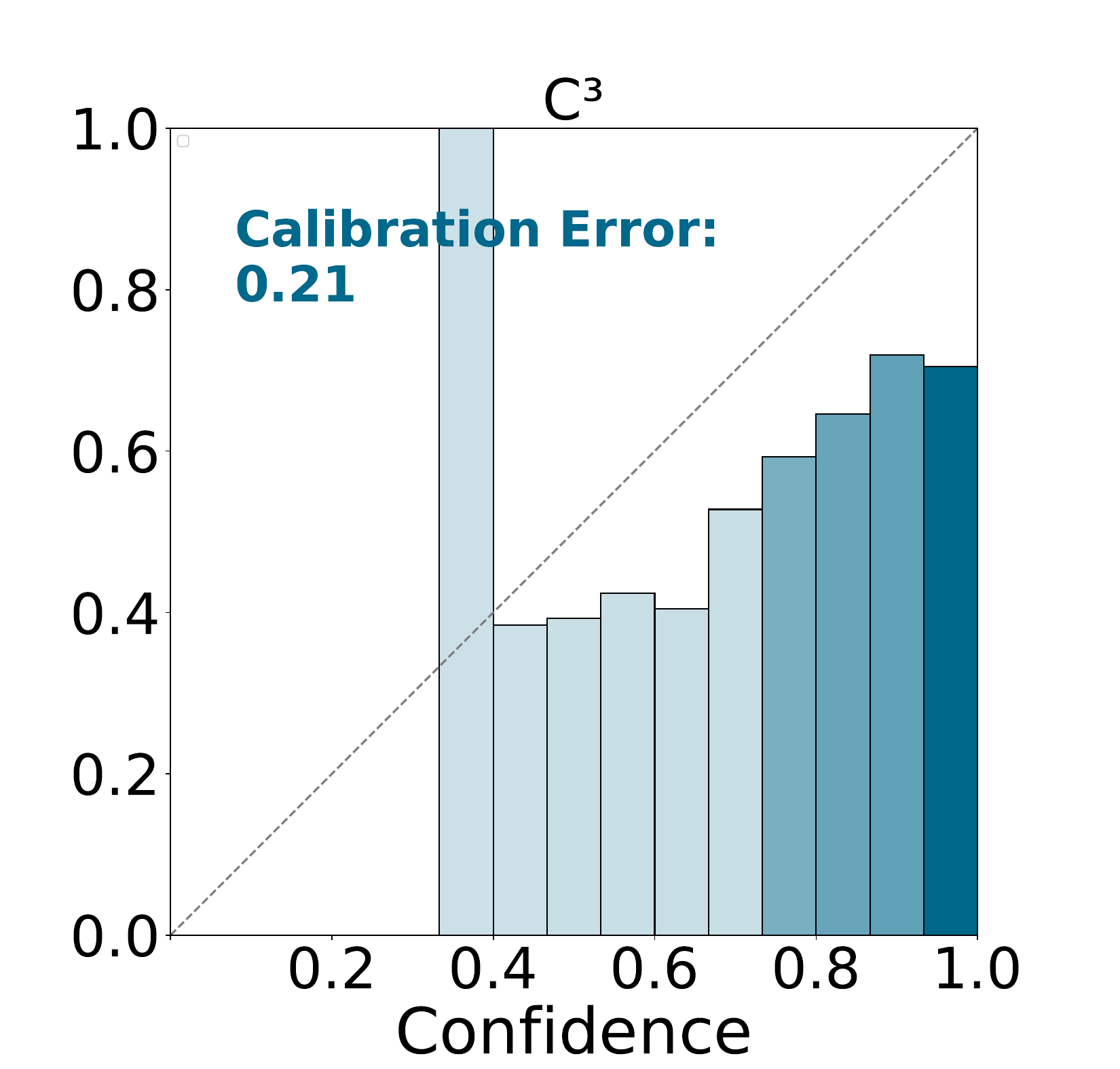}
      \caption{ECE comparison between Cascade and {\name} in Swahili language.}
      \vspace{20cm}
      \label{fig:sw_ece}
\end{figure}

We report the ECE figures for comparison between the Cascade (in orange) and {\name} (in blue) on Swahili and English in Figure \ref{fig:sw_ece} and Figure \ref{fig:en_ece}.

\vfill

\section{Experiment}
\subsection{Implementation Details}

In this section, we introduce the implemtation details of our {\name} method in two different settings: language model cascade (fine-tuning) and large language model cascade (zero/few-shot learning).

\noindent\textbf{Language Model Cascade} 
We initialize four encoder models from pretrained multilingual DistilBERT, multilingual BERT base, XLM-RoBERTa base, and XLM-RoBERTa large respectively. Each pretrained model is taken from the checkpoint published on Huggingface model hub. We then train each model independently on the English split in the training set for \{4, 4, 5, 5\} epoches with batch sizes of \{32, 32, 16, 16\} and learning rates of \{3e-5, 2e-5, 1e-5, 1e-5\} on 4 Nvidia A6000 GPUs.

\noindent \textbf{Large Language Model Cascade}
We initialize three casual language models from pretrained Llama-2-7b-chat-hf, Llama-2-13b-chat-hf, and Llama-2-70b-chat-hf published on Hugginface model hub. Without any training, we apply few-shot learning by giving the model three examples from the English split in the prompt.

\subsection{Datasets}

\subsection{Case Study}